\documentclass{article}

    \usepackage[preprint,nonatbib]{neurips_2024}

\usepackage[colorlinks=true, linkcolor=blue, citecolor=green, urlcolor=cyan]{hyperref}

\usepackage[utf8]{inputenc} 
\usepackage[T1]{fontenc}    
\usepackage{url}            
\usepackage{booktabs}       
\usepackage{amsfonts}       
\usepackage{nicefrac}       
\usepackage{microtype}      

\usepackage[table]{xcolor}  

\usepackage{amsmath}

\usepackage{algorithm}
\usepackage{algpseudocode}
\usepackage{changepage}
\usepackage{tocloft}
\usepackage{minitoc}

\usepackage[most]{tcolorbox}

\usepackage[utf8]{inputenc} 
\usepackage[T1]{fontenc}    

\usepackage{lipsum}

\usepackage{booktabs}

\usepackage{amssymb}
\usepackage{pifont}

\usepackage{graphicx}
\usepackage{hyperref}
\usepackage{listings}
\usepackage{enumitem}
\usepackage{multirow}

\usepackage{subcaption}
\usepackage{CJKutf8}

\usepackage{etoolbox,siunitx,booktabs,threeparttable,multirow,array,graphicx}
\usepackage[normalem]{ulem}
\robustify\bfseries
\robustify\uline

\usepackage{tabularx} %

\usepackage{eso-pic}
\usepackage{transparent}

\newcommand{\methodname}{\textsc{AugCon}}
\newcommand{\datasetname}{\textit{DailyM}}
\newcommand{\sftdata}{\textit{DailyM-SFT}}
\newcommand{\modelname}{Qwen-DailyM-32B}
\newcommand{\githublink}{\url{https://github.com/quanshr/AugCon}}

\title{Automatically Generating Numerous Context-Driven SFT Data for LLMs across Diverse Granularity}

\author{Shanghaoran Quan
   \\
  Beihang University \\
  \texttt{shrquan@buaa.edu.cn} \\}

\begin{document}

\maketitle

\begin{abstract}

Constructing high-quality query-response pairs from custom corpus is crucial for supervised fine-tuning (SFT) large language models (LLMs) in many applications, like creating domain-specific AI assistants or roleplaying agents. However, sourcing this data through human annotation is costly, and existing automated methods often fail to capture the diverse range of contextual granularity and tend to produce homogeneous data. To tackle these issues, we introduce a novel method named \methodname{}, capable of \underline{\textbf{au}}tomatically \underline{\textbf{g}}enerating \underline{\textbf{con}}text-driven SFT data across multiple levels of granularity with high diversity, quality and fidelity. \methodname{} begins by generating queries using the Context-Split-Tree (CST), an innovative approach for recursively deriving queries and splitting context to cover full granularity. Then, we train a scorer through contrastive learning to collaborate with CST to rank and refine queries. Finally, a synergistic integration of self-alignment and self-improving is introduced to obtain high-fidelity responses. 

Extensive experiments are conducted incorporating both human and automatic evaluations, encompassing a test scenario and four widely-used benchmarks in English and Chinese. The results highlight the significant advantages of \methodname{} in producing high diversity, quality, and fidelity SFT data against several state-of-the-art methods. All of our code, dataset, and fine-tuned model will be available at: \githublink{}.

\end{abstract}

\addtocontents{toc}{\protect\setcounter{tocdepth}{-1}}

\section{Introduction}

With the rise of impressive capabilities of large language models (LLMs), a variety of custom LLM-based AI assistants have been introduced~\cite{cheng2023adapting, chen2023disc, wu2023bloomberggpt, han2023medalpaca, liu2023chipnemo}. By incorporating specialized knowledge into LLMs, these custom models have been shown to outperform their general-purpose counterparts in their respective areas. These models can be developed through two strategies: building them from scratch~\cite{yang2023mindllm,liu2023context,korbak2023pretraining,dong2019unified} or adapting existing general LLMs through supervised fine-tuning~(SFT)~\cite{shi2023context,zaiem2023fine,luo2024taiyi,lin2024data}, with the latter approach often favored for its efficiency and the foundational advantages offered by the general LLMs~\cite{jiang2024improving,dong2023abilities,hu2021lora,wang2023self2}. 

Directly supervised fine-tuning on the raw, custom corpus, also known as domain-adaptive pre-training (DAPT)~\cite{gururangan2020don}, has proven beneficial~\cite{bayer2024cysecbert,krieger2022domain} but revealed to be insufficient and may impair prompting ability on domain-specific tasks~\cite{liu2023tailoring,pal2024domain}. To better leverage the privatized knowledge and customize the outputs of LLMs, supervised fine-tuning using custom query-response pairs has become common practice~\cite{shaikh2024rehearsal,chang2023llm4ts,bi2024deepseek,durante2024interactive,xu2023wizardlm}. However, sourcing these pairs through human annotation is very costly and can't generate at scale. Recent studies have explored automated methods for creating these pairs from custom corpus. AdaptLLM~\cite{cheng2023adapting}, for instance, has used regex-based patterns to generate query-response pairs, but this approach tends to produce a limited variety of SFT data, which may not significantly enhance prompting capabilities and risks overfitting due to the narrow range of query types predefined. ETRC~\cite{jiang2024improving} and Context-Instruct~\cite{wang2023rolellm} improved this by employing delicately designed prompts to generate queries from context using an LLM. However, those existing methods using the same workflow repeatedly on the same context tend to produce redundant queries without adequately covering the entire context at various levels of granularity. To automatically construct synthetic custom SFT data incorporating a wide range of contextual \textbf{granularity} (queries range from detailed questions to macro topics) with high \textbf{diversity} (queries need to be diversified to cover as much as possible the provided corpus), \textbf{quality} (responses are correct and efficient in answering the queries), and \textbf{fidelity} (data needs to follow human values and conform to predetermined tone and formats) still remain challenges.

To address these challenges, we propose \methodname{}, which automatically generates multi-granularity context-driven SFT data for LLMs at scale with high diversity, quality, and fidelity. \methodname{} performs the following three essential steps: 

\begin{figure}[t]
    \centering
    \includegraphics[width=0.9\linewidth]{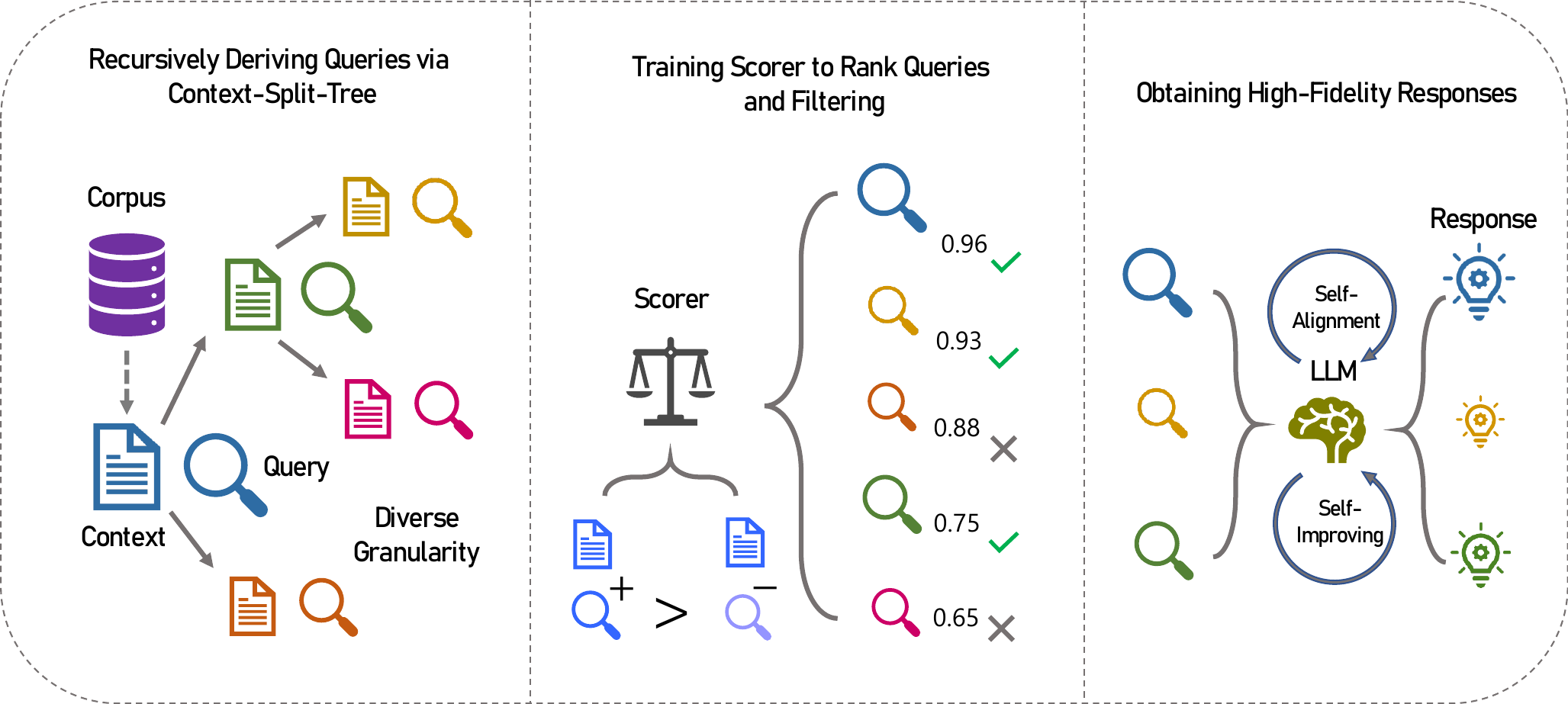}
    \caption{An overview of the proposed \methodname{}. }
    \label{fig:framework}
    \vspace{-10px}
\end{figure}


\begin{enumerate}[leftmargin=*]

    \item \textbf{Recursively Deriving Queries via Context-Split-Tree:} Considering that it is difficult for any predetermined prompts to generate multi-granularity queries from the same context, we propose a novel method called Context-Split-Tree (CST). Starting from a context (which is a continuous text chunk extracted from the corpus), we use an LLM to derive a query from it. At the same time, we ask the LLM to split this context into two contexts that are as independent as possible. Each context will recursively continue to derive queries and splits until it cannot be further divided. At the end, we will obtain a binary tree rooted in the initial context, and each node represents a context and contains a query that matches the granularity of it. 
    \item \textbf{Training Scorer to Rank Queries and Filtering:} To further ensure the quality and diversity of the queries, we use contrastive learning to train a scorer to evaluate the query by taking the obtained queries as positive examples and manipulating the prompt in Step 1 (\textit{e.g.}, by using suboptimal instruction or attaching fewer few-shot examples) to generate negative examples. Then, we sort the derived queries under the same context using the scorer and only retain the queries that get high scores and the diversity evaluated by ROUGE-L reaching a specific threshold. To ensure high quality and high diversity of queries while reaching the certain quantity requirements, the filtering stage will be iterated with CST until the requirements are met.
    \item \textbf{Obtaining High-Fidelity Responses:} Inspired by the significant impact principles~\cite{sun2024principle,sun2023salmon} have on LLMs, we employ a principle-driven self-alignment approach to guide the LLM in producing high-fidelity responses to filtered queries and their respective contexts. To enhance the quality of the generated answers further, we apply random search and conduct the LLM to self-evaluate its responses and discover the best in-context learning (ICL) examples from those annotated by humans. Ultimately, all context, ICL examples, and principles are discarded, leaving only the query-response pairs to supervised fine-tune the LLM.

\end{enumerate}

The entire process only requires a handful of few-shot CST examples, alignment principles, and query response examples. We can also achieve impressive results by just utilizing the open-source model, which will later be fine-tuned with synthetic data, eliminating the necessity of distilling more powerful LLMs like ChatGPT.

To assess the efficacy of our approach, we meticulously construct a test scenario and carefully assemble a dataset consisting of high-quality Chinese magazine articles centered around daily topics, along with corresponding test queries. Human evaluation demonstrates that our method excels in generating queries of superior quality and in enhancing the performance of fine-tuned models. Additionally, automatic evaluations conducted on four popularly used English benchmarks with relevant metrics further highlight the significant advantages our method holds in capturing contextual knowledge when compared to other state-of-the-art context-driven SFT data generation approaches.

Specifically, the contributions of our work lie on: 

\begin{itemize}

\item We propose \methodname{}, which can automatically generate multi-granularity context-driven SFT data from the corpus for LLMs at scale with high diversity, quality, and fidelity, providing the solution to a problem worth studying. 

\item Our ideas of deriving queries via CST, training the scorer using contrastive learning to collaborate with the generation process to refine data, and synergistic integrating self-alignment and self-improving to obtain high-fidelity responses, are very novel and may inspire further works.

\item Extensive experiments incorporating both human and automatic evaluations, encompassing the test scenario and widely-used benchmarks in English and Chinese compared with other state-of-the-art methods demonstrate the effectiveness and advantages of \methodname{}.

\item To boost the academy and for others to generate high-diversity SFT data on their own corpus without effort, we open-source all of our code, dataset, and fine-tuned model at: \githublink{}.

\end{itemize}

\section{Our Method: \methodname{}}
\label{sec:method}

In this section, we delve into the details of our proposed \methodname{}. A more comprehensive explanation is presented in Appendix~\ref{appx:method-details} given the space limitation. Additionally, to facilitate a direct understanding of how each step functions collectively, a case demonstration is provided in Appendix~\ref{appx:case-demo}.

\subsection{Preliminary}


We have a raw custom corpus $\mathcal{C} = \{C_1, C_2, \dots, C_n\}$ with each context $C_i$ represents a continuous text chunk extracted from corpus $\mathcal{C}$, the instruct prompt $I_{\text{CST}}$ and few-shot examples $E_{\text{CST}}$ for Context-Split-Tree and $I_R$ and $E_R$ for answering the queries, and several response principles $\mathcal{P}$ representing the human demands on responses when answering questions~\footnote{In this work, we use query-response and question-answer interchangeably.}. The $E_R$ are context-query-response triplets and will follow the response principles, represented as $E_R \sim \mathcal{P}$.

Our task is to generate numerous SFT query response pairs $\mathcal{D} = \{(q_{i,j}, r_{i,j})\}$ that each pair derives from either whole or part of context $C_i$. The derived triplet $(C, q, r)$ should also follow the response principles $\mathcal{P}$, and the generated $\mathcal{D}$ is expected to have high diversity, quality, and fidelity.

\subsection{Recursively Deriving Queries via Context-Split-Tree}
\label{sec:CST}

This step is to derive context-query pairs $(C, q)$ from the given corpus $\mathcal{C}$. Previous approaches applied regex-based or predetermined prompts for query generation, which often led to queries that were relatively monotonous in structure and granularity. We believe that this type of approach did not fully exploit the context, leading to queries incapable of effectively provoking the model's capability to comprehend and differentiate between various levels of detail within the context, resulting in suboptimal outcomes.

To address this issue, we propose a very novel and effective method called Context-Split-Tree (CST), with the pseudocode shown in Algorithm~\ref{algo:CST}. CST starts with an entire context $C$, with each attached with the instruct prompt $I_{\text{CST}}$ and few-shot examples $E_{\text{CST}}$ to call an LLM to generate a query $q$ deriving from the entire context. At the same time, we ask the LLM to semantically divide the context into two child contexts $C_1$ and $C_2$, and the instruct prompt is designed with hints to let the LLM polish the two split contexts to make them as independent as possible and collectively encompass the entirety of the original context. Each child context will continue to recursively derive query and split until reaching a point where one of its split child context lengths is not less than itself or the length falls below a predetermined threshold~$\lambda$. At this point, we consider it to have been split into the minimum granularity and cannot be further divided. Upon the completion of this recursive process, a binary tree structure is formed, with the initial context at the root, and each node representing a context along with its corresponding query tailored to its specific granularity. We collect data from all nodes as the outcome of this step. The detailed prompt templates and several case demonstrations are attached in Appendix~\ref{appx:CST-prompt-template} and \ref{appx:case-demo}, respectively.

\begin{center}
\begin{minipage}{0.8\textwidth}
\begin{algorithm}[H]
\caption{Context Split Tree}\label{algo:CST}
\textbf{Input:} A corpus $\mathcal{C}$, CST prompt instruction $I_{\text{CST}}$, CST few-shot examples $E_{\text{CST}}$\\
\textbf{Output:} Query dataset $Data$ comprises of split context and derived query pairs
\begin{algorithmic}[1]
\Function{ContextSplitTree}{$C, Data$}
    \If{$len(C) < \lambda$}
        \State \Return{} \Comment{Below the minimum granularity}
    \EndIf

    \State Call LLM to get $C_1, C_2, q \gets $ \Call{LLM}{$I_{\text{CST}}, E_{\text{CST}}, C$}
    \State Append $(C, q)$ to $Data$
    \If{$len(C_1) \geq len(C)$ \textbf{or} $len(C_2) \geq len(C)$ \textbf{or} ROUGE-L[P] $< 0.7$}
        \State \Return{} \Comment{The signs of hallucinations}
    \EndIf
    \State \Call{ContextSplitTree}{$C_1, Data$} \Comment{Recursive calling}  
    \State \Call{ContextSplitTree}{$C_2, Data$}
\EndFunction
\State
\State Initialize $Data \gets \text{empty list}$
\For{each extracted context $C \in \mathcal{C}$} \Comment{Extraction method is in Appx~\ref{appx:CST}}
    \State \Call{ContextSplitTree}{$C, Data$}
\EndFor
\State \Return{$Data$}
\end{algorithmic}
\end{algorithm}
\end{minipage}
\end{center}

The minimum length threshold~$\lambda$ and the initial context length~$l$ are like the lower bound and upper bound to control the granularity distribution of generated questions. One can easily adjust the overall average granularity of generated queries by adjusting the length threshold. Similarly, if we seek to address more global questions, we can do it by simply increasing the initial context length, as long as the model's context window permits. One beneficial property of CST is that the number of questions ultimately generated will maintain a linear relationship with the length of the initial text provided (proof can be found in Appendix~\ref{appx:CST-proof}). This ensures that adjusting the length of the segmented contexts in the corpus does not lead to significant fluctuations in the total number of queries obtained, but rather merely shifts the distribution of query granularity. By employing CST, we can produce queries that span across different levels of details in the context, and these queries naturally have little redundancy or repetition, enabling more efficient use of the context information and stimulating the model’s capability to comprehend and grasp the context in different granularities. Moreover, another benefit of CST is that the derived queries just match the split context, making the later generated response to these queries more accurate and pertinent with less unrelated information.

\subsection{Training Scorer to Rank Queries and Filtering}
\label{sec:step2}

To further enhance the quality and diversity of the generated data, we introduce an effective ranking and filtering strategy collaborating with CST. Previous works have attempted to filter training data via heuristic algorithms, such as filtering out queries that are too long or too short~\cite{wang2023self}. Other works that are more relevant to us attempt to train scorers to judge the complexity and quality of question-response pairs~\cite{liu2023makes}, but they need to have a step of distillation on stronger LLM APIs like ChatGPT, and their training methods are less effective. For example, they put a series of responses and ask for direct ranking, suffering from the positional bias~\cite{liu2024lost} in LLMs, or ask LLMs to directly assign a scalar score to a response, which is unstable. In this work, we apply contrastive learning to train a scorer to judge the degree of adherence to instruct prompt and few-shot examples, which is data-efficient and can achieve effective performance without the need for stronger LLMs.

The structure of our scorer is obtained by adding a linear head after the base model to map the last hidden state to a one-dimensional space. We take context-query pairs as inputs, applying scorer~$Sc$ to yield a scalar score $s = Sc(C, q)$. We use the context query pairs obtained from Step 1 as positive samples: $q^+ =$ \Call{LLM}{$I_{\text{CST}}, E_{\text{CST}}, C$}, and obtain negative samples by manipulating the instruct prompt (use suboptimal instructions): $q^- =$ \Call{LLM}{${I_{\text{CST}}}^-, {E_{\text{CST}}}, C$}, few-shot examples (reduce ICL examples count): $q^- =$ \Call{LLM}{${I_{\text{CST}}}, {E_{\text{CST}}}^-, C$} or both of them: $q^- =$ \Call{LLM}{${I_{\text{CST}}}^-, {E_{\text{CST}}}^-, C$}. The details of constructing positive-negative pairs are presented in Appendix~\ref{appx:step2-training}. Note that we do not generate all corresponding negative examples for positive data for training scorer, but rather randomly select a very small number of samples (\textit{e.g.} only $500$ pairs for each negative types in our implementation) to form the training set~$D_{Sc}$. Then, the loss function of scorer can be represented as: 

\begin{equation}
    \mathcal{L} = - \mathbb{E}_{(C,q^+,q^-) \sim D_{Sc}}[\log(\sigma(Sc(C,q^+)-Sc(C,q^-))))]
\end{equation}

We use the trained scorer applied on all the context query pairs obtained in Step 1 to get their scores. For each root context, we rank all queries from its CST in descending order of scores. Then, we start with an empty set and add one training query each time, only if the current query has a ROUGE-L precision score of less than $0.7$ compared to any previously added queries. We will stop adding as the count reaches the limit. Each context will form such a set, and ultimately, we consolidate and retain the training data from all the sets. Through this approach, we can obtain diverse data and easily control the quantity, for it makes it possible to apply multi-times CST in the same context and filter the repeated one. The details of how the filtering pipeline cooperates with CST to improve the quality and diversity of queries are expatiated in Appendix~\ref{appx:step2-filter}.

\subsection{Obtaining High-Fidelity Responses}
 

Inspired by the significant impact principles~\cite{sun2024principle,sun2023salmon} have on LLMs, this principle-driven self-alignment step begins by appending the context and a set of helpful, ethical, and reliable principles to the LLM. These principles are meticulously crafted to ensure the LLM's outputs are closely aligned with human preferences or mimic certain response tones. Before initiating the response generation, we deploy a self-improving pipeline that makes the LLM self-evaluate its response and sift through the entire set of human-annotated Q\&A pairs $E_R$, where random search is applied to find the most fitting few-shot examples to help LLM generate high-fidelity responses under the predetermined principles, denoted as ${E_R}'$. The detailed implementation is shown in Appendix~\ref{appx:step3}.

The innovative synergistic integration of the principle-driven methodology with self-improving effectively improves the fidelity of generated responses. Following this, we execute \Call{LLM}{$I_R, {E_R}', C, q$} to elicit each response $r$, ensuring that each response is not only high in quality but also in alignment with our established principles. Notably, due to the precise matching of each query with its context’s granularity within the CST framework, the LLM can effortlessly provide accurate and pertinent responses to the queries.



After obtaining all generated data, we prune all context, instruction, and response principles and only retain synthetic query response pairs as SFT data. This approach allows the fine-tuned LLM to potentially learn the methods and nuances of responding to queries in a manner that naturally aligns with human expectations, enabling the LLM to directly generate responses that are well-aligned with reliable principles and optimal ICL exemplars across a wide range of queries. It’s important to note that the fine-tuned LLM can generate high-quality responses without the need to explicitly reference the principles set and ICL exemplars.

\section{Evaluations}
\label{sec:evaluations}

To validate the effectiveness of the proposed method, we apply human evaluation on a test scenario in Section~\ref{sec:human-evaluation} and conduct automatic evaluations on four popular and widely used benchmarks in Section~\ref{sec:automatic-evaluation}. The set of contexts, base language models, and quantity of retained query-response pairs are maintained the same (if applicable) on both the baselines and our method to ensure a fair comparison. To provide a more thorough evaluation of our method, we present extensive experiments evaluated from various perspectives in Appendix~\ref{appx:addtional-experiments}.

\subsection{Baselines}

To demonstrate the advantages of our method, we meticulously collect the following relevant baselines from a wide range of research, with the implementation details presented in Appendix~\ref{appx:implementation-details}.

(1) \textbf{Chat Model}~\cite{bai2023qwen,touvron2023llama2} applies instruction tuning and alignment tuning after pre-training. We utilize it both as the basic baseline and as the fundamental model for calling and fine-tuning across all other baselines and our methods for fair comparison.

(2) \textbf{DAPT}~\cite{gururangan2020don} continuously pre-trains directly on the raw custom corpus to adapt and grasp domain-specific knowledge. 

(3) \textbf{AdaptLLM}~\cite{cheng2023adapting} builds SFT samples by converting the raw corpora into reading comprehension tasks via
regex-based mining patterns. Tasks they design include summarization, word-to-text, natural language inference, commonsense reasoning, and paragraph detection. 

(4) \textbf{ETRC}~\cite{jiang2024improving} derives question-answer pairs from extracted contexts with an LLM and augments data by ensembling contexts and their corresponding question-answer pairs with a length-based clustering algorithm. their corresponding question-answer pairs with a length-based clustering algorithm. 

(5) \textbf{Context-Instruct}~\cite{wang2023rolellm} is a context-driven instruction generation method that contains three parts: 1) partition text into manageable segments, 2) use an LLM to generate question, response, and confidence score triplets based on the segments, and 3) apply confidence-score-based filtering and deduplication to ensure data quality and diversity.

\subsection{Human Evaluation}
\label{sec:human-evaluation}

To ensure the efficacy and reliability of our methodology, we build a test scenario and conduct a comprehensive human evaluation. This evaluation compares our constructed data and the fine-tuned model against various baselines, aiming to provide a rigorous assessment of performance enhancement and validation of our techniques. The human evaluation protocol is designed to provide a nuanced understanding of the improvements our method offers, ensuring that the enhancements are not just statistically significant, but also meaningful and perceptible to the users. 


Specifically, we meticulously curate a corpus dataset, referred to as the \datasetname{} dataset, which consists of $1,000$ articles carefully selected from a variety of high-quality Chinese magazines closely related to daily life. These articles extensively cover issues of widespread public concern such as basic livelihood, politics, economics, and law, with each article containing approximately $4,000$ Chinese characters. Then, \textbf{we test how well our method and baselines build an AI chat assistant specialized in this daily concern corpus.} We apply our method on \datasetname{} to generated SFT data called \sftdata{} and use these data to fine-tune Qwen1.5-32B-Chat~\cite{bai2023qwen} to get fine-tuned model \modelname{}. To further test our method, we conduct annotators to write a total of $1,000$ queries they are interested in related to these articles, forming the \datasetname{} test set. To facilitate further research and development within the research community, we plan to make our \datasetname{}, the constructed SFT data \sftdata{}, and the fine-tuned model \modelname{} all open-sourced.

\subsubsection{Metrics}
\label{sec:human-eval-metrics}

In our comprehensive evaluation framework, we assess both the generated queries and the outputs under the \datasetname{} test set of the fine-tuned models. This dual approach ensures a holistic understanding of the method's performance, encompassing the generation of realistic, diverse queries and the quality of the responses provided by the fine-tuned models. The evaluation metrics have been tailored to address the specific characteristics and objectives users are concerned about in real scenarios.

For generated queries:

\begin{enumerate}[leftmargin=*]
    \item \textbf{Realism}: This metric evaluates how closely the generated queries resemble those that would be posed by users, and whether they curious or willing to ask this question. Evaluators will consider the naturalness and authenticity of the queries, scoring them on a scale from 1 (completely artificial) to 5 (indistinguishable from human-created).
    \item \textbf{Diversity}: Reflecting on the range of topics and the variety of the generated queries. A score from 1 (very monotonous) to 5 (highly diverse) will be assigned, with higher scores indicating a wide spectrum of query types and topics, including various levels of granularity.
\end{enumerate}

For fine-tuned models' outputs:

\begin{enumerate}[leftmargin=*]
    \item \textbf{Relevance}: This assesses how relevant the model's responses are to the test queries. It is crucial that the responses accurately address the queries' intent, providing meaningful and contextually appropriate information. Scores will range from 1 to 5, with 5 being the most relevant.
    \item \textbf{Accuracy}: This metric measures the factual correctness of the responses, with a score from 1 (many hallucinations) to 5 (completely accurate). Accuracy is paramount, and evaluators are provided with relevant context and external tools like search engines to support their judgments.
    \item \textbf{Satisfaction}: Reflecting the degree of satisfaction, this general metric allows evaluators to rate their total satisfaction with the responses, serving as an overall assessment. The scoring will be from 1 (highly dissatisfied) to 5 (highly satisfied).
\end{enumerate}

For both generated queries and model outputs, evaluators are provided with detailed scoring rubrics and examples to promote consistency in evaluation. The queries and outputs will be reviewed by multiple independent evaluators to ensure a balanced and objective assessment, with average scores calculated for each metric to determine the overall performance. This comprehensive human evaluation metrics approach is designed to rigorously assess the effectiveness and applicability of our method in generating multi-granularity and provoking queries and producing high quality and fidelity responses. We provide the detailed evaluation guidance in Appendix~\ref{appx:guidance}.

\subsubsection{Results}

For all our baselines and the proposed \methodname{}, we employ Qwen1.5-32B-Chat~\cite{bai2023qwen}, a popular open-source LLM proficient in Chinese, as the fundamental model for calling and conducting fine-tuning for evaluations. For methods such as AdaptLLM, ETRC, Context-Instruct, and our \methodname{} which generate query-response pairs based on context, we adhere to a standard where every $35$ Chinese characters derive one query-response pair to ensure a fair comparison. We limit the number of generated entries to the same in the comparison because we find that all methods spend much more time on final fine-tuning process compared to the previous generation. Meanwhile, we also provide a GPT-4 judge in Appendix~\ref{appx:gpt-4-judge}, a computation experiment in Appendix~\ref{appx:computation-experiment}, and a training phase experiment in Appendix~\ref{appx:training-phase} for a more comprehensive comparison.

\begin{figure}[h]
    \centering
    \includegraphics[width=0.8\linewidth]{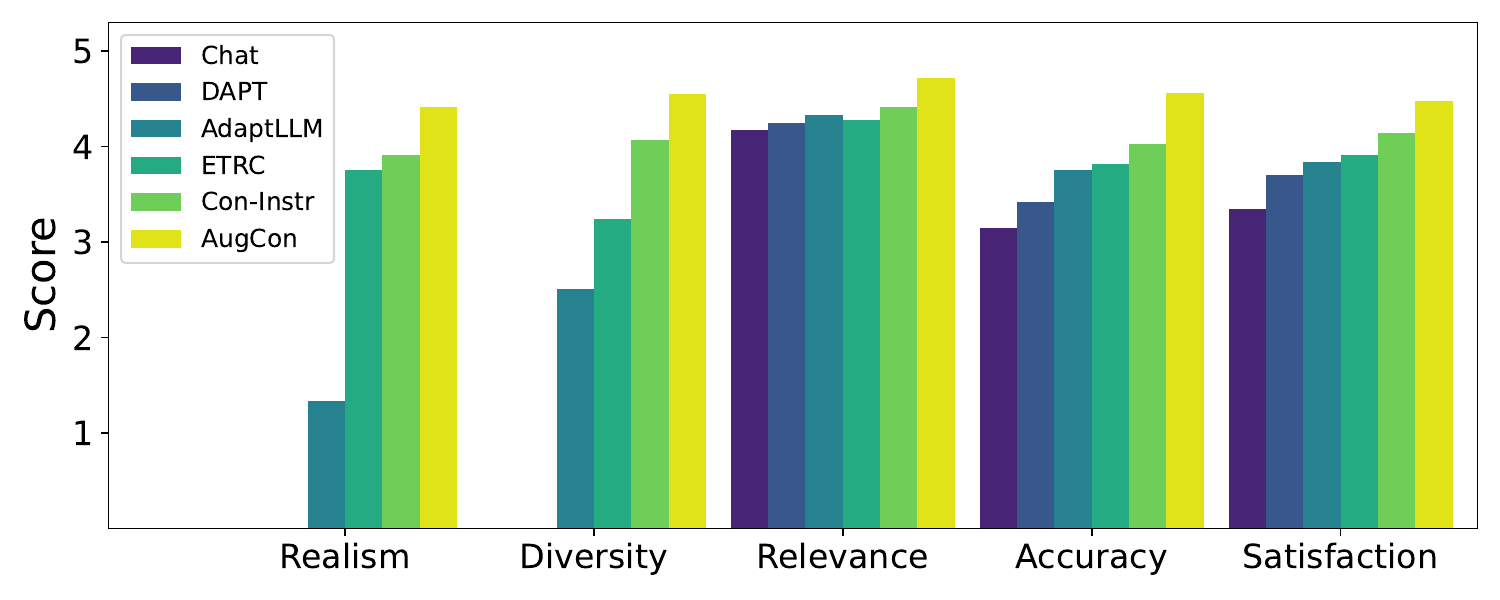}
    \caption{The results of human evaluation on \datasetname{}. Query metrics are not applicable for the base chat model and DAPT so we don't show them.}
    \label{fig:result-human-eval}
\end{figure}

Figure~\ref{fig:result-human-eval} presents the results of a human evaluation on the \datasetname{} test set. The results demonstrate that \methodname{} consistently surpasses the baseline methods across all evaluation metrics. Specifically, the superior performance in terms of query realism and diversity underscores our method's ability to produce human-like and high-diversity queries. Since our CST and filtering process effectively gain multi-granularity queries that are more effective in covering all granularity levels of context, the derived data will extract more useful knowledge from the corpus. Furthermore, the impressive performance in judging relevance, accuracy, and satisfaction in responses from fine-tuned models further validates that our method's high-quality and diverse queries, coupled with high-fidelity responses, can indeed enhance the performance of subsequently fine-tuned models and achieve higher satisfaction scores from humans. This suggests that \methodname{} is particularly adept at constructing high-quality supervised fine-tuning data for LLMs from a given corpus.


\subsection{Automatic Evaluation}
\label{sec:automatic-evaluation}

To objectively assess the impact of our approach, we conduct automatic evaluations on four widely used benchmarks, employing relevant metrics to facilitate a direct comparison between models fine-tuned with our generated data and established baselines. This concise evaluation method provides clear, objective insights into the efficacy of our data construction paradigm, highlighting the potential advantages of our \methodname{} over existing methods.

\subsubsection{Benchmarks}

To automatically evaluate our method and baselines, we meticulously collect a range of short-form and long-form question-answering benchmarks that have corpus or contexts for reference (we re-compile all reference contexts as the corpus). All baselines and our \methodname{} are applied on the same provided corpus and test on the test set. All the public links to these benchmarks are listed in Appendix Table~\ref{tab:link-dataset}.

(1) \textbf{SQuAD1.1}~\cite{rajpurkar2016squad} is a reading comprehension dataset, consisting of questions posed by crowdworkers on a set of Wikipedia articles, where the answer to every question is a segment of text, or span, from the corresponding reading passage. SQuAD1.1 contains $100,000+$ question-answer pairs on $500+$ articles.

(2) \textbf{TriviaQA}~\cite{joshi2017triviaqa} includes 95K question-answer pairs authored by trivia enthusiasts and independently gathered evidence documents, six per question on average, that provide high quality distant supervision for answering the questions.

(3) \textbf{DROP}~\cite{dua2019drop} is a crowdsourced, adversarially-created, 96K-question benchmark, in which a system must resolve references in a question, perhaps to multiple input positions, and perform discrete operations over them.

(4) \textbf{WebGLM-QA}~\cite{liu2023webglm} is the data used to train the WebGLM generator module and an LLM bootstrapped quoted and long-formed QA dataset via in-context learning and corresponding strategies to clean and refine, with 45K high-quality filtered and 83K unfiltered samples.


\subsubsection{Metrics}

For datasets featuring short-form responses (applied to the SQuAD1.1, TriviaQA, and DROP datasets), we measure the model's performance using accuracy~(Acc). A response is considered correct if and only if it matches any of the possible answers. For datasets with long-form responses (applied to the WebGLM-QA dataset), we employ BERTScore~(BS)~\cite{zhang2019bertscore} (we use Roberta-Large~\cite{liu2019roberta} for calculation) to evaluate the semantic similarity between the generated outputs and the reference responses. 

\subsubsection{Results}

\begin{table*}[h]
\centering
\begin{tabular}{lrrrr}\toprule
&\multicolumn{3}{c}{Short-form (Acc)} & \multicolumn{1}{c}{Long-form (BS)}\\
Method&SQuAD1.1&TriviaQA&DROP&WebGLM-QA\\
\midrule

Llama3-c$_{\textsc{70b}}$&
    $0.212 \scriptstyle \pm 0.004$&
    $0.723 \scriptstyle \pm 0.003$&
    $0.220 \scriptstyle \pm 0.004$&
    $0.837 \scriptstyle \pm 0.002$ \\
DAPT&
    $0.258 \scriptstyle \pm 0.004$&
    $0.767 \scriptstyle \pm 0.003$&
    $0.266 \scriptstyle \pm 0.004$&
    $0.851 \scriptstyle \pm 0.002$ \\
AdaptLLM&
    $0.273 \scriptstyle \pm 0.003$&
    $0.791 \scriptstyle \pm 0.004$&
    $0.284 \scriptstyle \pm 0.003$&
    $0.842 \scriptstyle \pm 0.001$ \\
ETRC&
    $0.301 \scriptstyle \pm 0.004$&
    $0.812 \scriptstyle \pm 0.003$&
    $0.326 \scriptstyle \pm 0.004$&
    $0.903 \scriptstyle \pm 0.001$ \\
Context-Instruct&
    $0.314 \scriptstyle \pm 0.003$&
    $0.825 \scriptstyle \pm 0.003$&
    $0.334 \scriptstyle \pm 0.003$&
    $0.885 \scriptstyle \pm 0.001$ \\
\methodname{}\textit{(Ours)}&
    $\mathbf{0.336} \scriptstyle \pm 0.004$&
    $\mathbf{0.849} \scriptstyle \pm 0.003$&
    $\mathbf{0.350} \scriptstyle \pm 0.003$&
    $\mathbf{0.924} \scriptstyle \pm 0.002$ \\
\bottomrule
\end{tabular}
\caption{The results of automatic evaluation on four benchmarks. We run 10 times for each test and report the mean value and standard deviation, with the best results shown in bold.}
\label{tab:result-auto-eval}
\end{table*}

We employ Llama3-70B-Instruct~\cite{llama3modelcard} as the fundamental model for calling and conducting fine-tuning for automatic evaluations for all our baselines and the proposed \methodname{}. The detailed results are shown in Table~\ref{tab:result-auto-eval}. The results illustrate that our proposed method consistently outperforms the established baselines across all four datasets. Specifically, when analyzing short-form datasets, it becomes evident that the data generated by \methodname{} surpasses the comparative methods in extracting pivotal information and knowledge from the corpus, thus improving the question-answering accuracy of fine-tuned models.
Meanwhile, the exceptional performance of \methodname{} on datasets emphasizing long-form responses showcases its proficiency in generating high-fidelity query-response pairs. This capability directly contributes to enhancing the effectiveness of chat models, enabling them to deliver more relevant, engaging, and contextually appropriate responses based on the given corpus. This, in turn, significantly improves the overall user experience by ensuring that interactions are not only informative but also closely aligned with the user's specific curiosities and requirements.

Furthermore, the consistency of \methodname{} in achieving top results across all four datasets, each with unique query patterns and focuses, speaks volumes about its versatility and adaptability. Such consistent performance across varied datasets underscores the robust generalization ability of our method, making it a highly effective tool for a broad spectrum of corpus types and catering to diverse user interests and inquiries. We also provide ablation studies and further analysis in Appendix~\ref{appx:ablation-study}.

\section{Related work}

\paragraph{Synthetic Data for Language Models} Due to the challenges of data scarcity~\cite{babbar2019data}, privacy concerns~\cite{abay2019privacy}, and the sheer cost of data collection and annotation~\cite{gilardi2023chatgpt}, synthetic data has emerged as a promising solution to build large, diverse, and high-quality datasets at scale~\cite{liu2024best}. One benefit of synthetic data is it can be tailored to specific requirements~\cite{cheng2023adapting,jiang2024improving,liu2023webglm}, with practical applications having been employed in various domains. WizardMath~\cite{luo2023wizardmath} leverages a series of operations to increase the complexity of questions and answers using GPT-3.5, while Reflexion~\cite{shinn2024reflexion} employs external or internally simulated linguistic feedback to improve the code reasoning capabilities of language models. Similarly, Toolformer~\cite{schick2024toolformer} learns to decide which APIs to call and what arguments to pass by training on template-generated data. In addition, synthesized data has been proven effective in mitigating hallucination~\cite{wei2023symbol,wei2023simple,jones2023teaching,tian2023fine} and aligning with shared human preferences and values~\cite{bai2022constitutional,sun2024principle,sun2023salmon,quan2024dmoerm,kim2023aligning}. While the generation of context-driven synthetic data has proven to be a powerful substitute for manual annotation, the challenge of ensuring high-quality synthetic data, which encompasses the complexity of queries~\cite{liu2023makes,li2024common,liu2024augmenting}, the diversity of semantics~\cite{ding2023enhancing,wang2023data,wang2024survey,lu2023instag,liu2024chatqa}, and the scale of the synthetic datasets~\cite{yuan2023scaling,gunasekar2023textbooks,li2023textbooks}, has been a consistent pursuit.

\paragraph{Context-Driven Synthetic Data} Numerous studies have developed techniques for creating synthetic data informed by contextual cues. UltraChat~\cite{ding2023enhancing} leverages user-specified topics and supplements these with existing textual material to craft instructional conversations aimed at enhancing chatbot performance. SPIN~\cite{chen2024self}, on the other hand, autonomously generates training data from its previous iterations, employing this approach to progressively refine its capabilities. RECOST~\cite{zhang2024recost} selects top-tier instructional content by incorporating external knowledge to assess synthesized examples using an in-context relative predictive entropy measure.
Additionally, various methods have been devised to extract character profiles and personas from collected books or scripts for the purpose of producing roleplaying dialogues~\cite{shao2023character,zhou2023characterglm,wang2023rolellm,li2024camel}, and several initiatives focus on mining domain-specific data from specialized corpora to construct domain-specific language models~\cite{cheng2023adapting,jiang2024improving,frey2023neural,deng2024k2,yue2023disc}.
While alternative approaches employ retrieval augmented generation (RAG)~\cite{ram2023context,borgeaud2022improving} or integrate auxiliary knowledge in vast context windows~\cite{xiong2023effective,an2024training}, issues like entity susceptibility~\cite{du2024context}, high inference computational demand~\cite{liu2022few,hao2022structured}, and alignment difficulties with formats and preferences~\cite{qi2023fine,mosbach2023few,addlesee2023multi} highlight the crucial role of context-driven SFT in effectively incorporating corpus knowledge internally.





\section{Conclusion}

In this work, we propose \methodname{}, a highly innovative and effective method to build custom AI assistants from the corpus by deriving SFT query-response pairs with diverse granularity. \methodname{} starts with query generation through the Context-Split-Tree (CST), an innovative approach for recursively deriving queries and splitting context to cover full granularity. We then employ contrastive learning to develop a scorer that works with CST to rank and refine queries. Finally, we introduce a synergistic integration of self-alignment and self-improving to obtain high-fidelity responses. We conduct extensive experiments on Qwen1.5-32B-Chat and Llama3-70B-Instruct models. The human evaluation on a test scenario and automatic evaluation on four benchmarks demonstrate the significant advantages of our method in producing high diversity, quality, and fidelity context-driven SFT data and improving the performance of custom fine-tuned models against existing methods.




\medskip

{
\small
\bibliographystyle{plain}
\bibliography{custom}
}


\newpage
\appendix




\renewcommand{\contentsname}{Appendix} 
\tableofcontents
\addtocontents{toc}{\protect\setcounter{tocdepth}{3}}
\newpage

\section{Limitations}
\label{appx:limitations}

While our research introduces an innovative approach to automatically generating multi-granularity SFT data for LLMs based on context, several limitations should be acknowledged. These limitations highlight areas for future research and potential improvements in our methodology.

\paragraph{Contextual Depth and Complexity}
Our method relies heavily on the context provided for SFT data construction. However, the complexity and the depth of the context can vary significantly, potentially impacting the quality and the relevance of the generated SFT data. In instances where the context is too narrow or lacks depth, the model may produce SFT data that is not sufficiently diverse or representative.

\paragraph{Model Bias and Sensitivity}
Like all machine learning models, our approach is subject to the biases inherent in the training data. Despite efforts to mitigate these biases, there may still be underlying biases that affect the SFT data generation process. Additionally, our method’s sensitivity to nuanced linguistic and cultural differences may not be fully addressed, possibly leading to the generation of data that might not be entirely appropriate or sensitive to all contexts.

\paragraph{Scalability and Computational Resources}
While our method is designed to automate the construction of SFT data for LLMs, the scalability of this approach can be constrained by available computational resources. The processing power required to analyze complex contexts and generate high-quality SFT data can be substantial, which may limit the applicability of our method in resource-constrained environments.

\paragraph{Generalization Across Different Languages and Domains}
Our initial experiments and results are promising but are primarily focused on specific languages and domains. The ability of our method to generalize across different languages, dialects, and specialized domains has not been thoroughly tested. This limitation suggests that the effectiveness of our approach may vary significantly when applied to less common languages or highly specialized fields.

\paragraph{Evaluation Metrics and Benchmarks}
The evaluation of the automatically constructed SFT data’s quality relies on metrics that may not fully capture the nuances of semantic fidelity and context appropriateness. The development of more refined evaluation benchmarks and metrics that can better assess the quality and utility of SFT data in training LLMs remains an area for future work.

\paragraph{Conclusion}
While our method presents a novel approach to enhancing the training of large language models through contextually generated SFT data with multi-granularity, these limitations underscore the importance of continued research and development. Addressing these challenges will be crucial for improving the efficacy and applicability of automated SFT data construction methods in the field of large language model.

\section{Boarder Impacts}
\label{appx:boarder-impacts}

The advent of large language models has significantly advanced the capabilities of artificial intelligence in understanding and generating human-like text. Our research presents a novel methodology for automatically generating SFT query response data pairs based on context with multi-granularity, a development that holds considerable implications for the field of AI and its applications across various domains. This section explores the broader impact of our work, encompassing both its potential benefits and challenges.

\paragraph{Enhancing Model Performance and Efficiency}

The automated construction of SFT datasets tailored to the context at hand has the potential to significantly improve the performance of LLMs. By providing high-quality, targeted training data, models can achieve better understanding and generate more accurate outputs, thereby increasing their utility in a wide range of applications. Furthermore, this method reduces the need for extensive manual dataset preparation, leading to more efficient model development.

\paragraph{Advancing Domain-Specific Applications}
Our method stands to greatly enhance the development of domain-specific applications, such as specialized question-answering assistants. By enabling the automatic generation of SFT datasets tailored to specific contexts, our approach facilitates the creation of LLMs that are not only more accurate but also more relevant for specialized fields such as medicine, law, and engineering. This could lead to significant improvements in professional assistance systems, offering experts timely and accurate information and potentially accelerating decision-making processes in critical situations.

\paragraph{Democratizing AI Development}
By automating the construction of SFT datasets, our methodology could democratize access to high-quality AI development. Smaller organizations or groups with limited resources might find it easier to develop powerful, context-specific AI tools without the need for extensive manual dataset curation. This democratization could accelerate innovation and competition, leading to a broader range of AI applications and services available to the public.

\paragraph{Educational Implications}
Our approach could also have profound implications for education. Customized LLMs can be developed to provide students with personalized learning assistants. These AI tutors could adapt to each student’s learning style and pace, offering explanations, supplementary information, or exercises based on the context of the student’s needs. Such personalized education could enhance learning outcomes and make education more accessible and effective for diverse learners.

\paragraph{Ethical and Societal Considerations}
However, the broader deployment of context-specific LLMs also raises important ethical and societal considerations. The accuracy and fairness of these models depend on the quality and diversity of the SFT data. There is a risk that biases present in the training data could be amplified, leading to unfair or harmful outcomes. It is crucial to develop methodologies for monitoring and mitigating bias in these models.

Moreover, the widespread use of powerful, domain-specific LLMs could have unforeseen impacts on employment, particularly in sectors where decision-making or informational roles are automated by AI. While these technologies can augment human capabilities, there is also a need for policies that address potential displacements and ensure that the benefits of AI advancements are broadly shared across society.


\paragraph{Conclusion}
The automatic construction of context-specific SFT datasets for LLMs as proposed in our research has the potential to significantly impact various sectors positively. However, it is imperative to navigate the ethical, societal, and environmental challenges associated with these advancements. By addressing these issues proactively, we can ensure that the benefits of AI are realized equitably and sustainably across society.

\section{Method Details}
\label{appx:method-details}

In this section, we delve into the detailed framework of our methods, providing a thorough examination of each step in practice. Accompanied by comprehensive prompt templates and detailed implementation specifics, we aim to offer a clear understanding of how our approach functions in action.

\subsection{Recursively Deriving Queries via Context-Split-Tree}
\label{appx:CST}

The Context-Split-Tree construction process prepares by dividing the corpus into short, consecutive text contexts, each with a limit of $500$ words. Gain inspiration from retrieval augmentation methods~\cite{sarthi2024raptor}, if a sentence surpasses the $500$-word threshold, we move the entire sentence to the next context, rather than cutting it mid-sentence. This approach maintains the contextual integrity and semantic consistency of the text in each context. After obtaining the extracted contexts, for each context, we construct a correspondence CST in the manner described previously in Section~\ref{sec:CST}. The specific templates and few-shot examples used are detailed in Appendix~\ref{appx:CST-prompt-template}.  

\subsubsection{Proof of Linear Relationship}
\label{appx:CST-proof}

A commendable property is that with the initial text length~$l$ provided, we can achieve multi-granularity effects through a linear quantity of generated questions. To be more precise, the number of questions generated will have a linear relationship with the number of minimum sentence units contained in the initial context. This allows us to generate different distributions of query granularity by simply adjusting the minimum sentence length $\lambda$ and the initial context length without worrying about significant fluctuations in the computation consumption or the overall number of queries obtained. We will prove this property in the following.

We represent context as a collection of sentences $C = \{S_1,S_2,\dots,S_n\}$. We assume that during the CST process, these sentences are the smallest units and will not be split internally or increased in quantity (they may be polished but do not change the essential semantics). Formally, we have the following assumption.

\paragraph{Assumption:} Given a context $C = \{S_1,S_2,\dots,S_n\} (n > 1)$, using LLM for a split will definitely yield one question $q$ and two child context $C_1$ and $C_2$ that satisfy that there exists an integer $ 1 \leq i < n$ such that $C_1 = \{S_1,\dots,S_i\}$ and $C_2=\{S_{i+1},\dots,S_n\}$. Specifically, when the context degrades to a single sentence, calling the LLM will generate a question and terminate.

Then, based on the preceding assumptions, we have the following proposition.

\paragraph{Proposition:} For any context $C = \{S_1, S_2, \dots, S_n\}$ containing $n$ sentences, where $n$ is an arbitrary positive integer, applying CST to it will ultimately generate $2n - 1$ questions.

\paragraph{Proof:} We will prove the proposition using the \textit{Second Principle of Mathematical Induction}.

\begin{enumerate}[leftmargin=*]

    \item First, for $n=1$, calling the LLM will generate a question and then terminate, which is consistent with the proposition. 

    \item Secondly, assume the conclusion holds for all \(n \leq k\) (\(k \geq 1\)). When \(n = k+1\), according to the assumption, calling the LLM with \(C = \{S_1, \ldots, S_{k+1}\}\) will generate a question \(q\), along with two sub-contexts \(C_1\) and \(C_2\), where there exists \(1 \leq i < k+1\) such that \(C_1 = \{S_1, \ldots, S_i\}\), and \(C_2 = \{S_{i+1}, \ldots, S_{k+1}\}\). The numbers of sentences in \(C_1\) and \(C_2\) are $i$ and $k+1-i$ and both strictly less than \(k+1\). By assumption, \(C_1\) will eventually generate \(2i-1\) questions, and \(C_2\) will generate \(2(k+1-i)-1 = 2(k-i)+1\) questions. Therefore, in total, \(C\) will generate \(1 + (2i-1) + (2(k-i)+1) = 2k + 1 = 2n - 1\) questions by the end. Thus, the proposition also holds for \(n = k +1\).

    \item Finally, combining 1 and 2 along with the \textit{Second Principle of Mathematical Induction}, it can be concluded that the proposition holds. 

\end{enumerate}

Therefore, a context containing \(n\) sentences will ultimately generate \(2n - 1\) questions, which establishes a linear relationship between the number of sentences in the context and the number of questions generated. \hfill $\square$

\subsubsection{Prompt Template \& Few-Shot Examples}
\label{appx:CST-prompt-template}

Our method is applicable across various languages, for which we provide the utilized prompt templates in both English and Chinese. 

For English version:




























\phantomsection
\label{CST-template-English}
\begin{tcolorbox}
[breakable,colback=white!95!gray,colframe=gray!50!black,rounded corners, title={Prompt Template for Context-Split-Tree with Instruction in English}]

Given an entire context as the Context, generate a Question about the entire context that users might be interested in, which answer should be able to be derived directly from the Context. Then, divide the entire context into two sub-contexts Context 1 and Context 2 based on their semantic content, making necessary adjustments within each sub-contexts to ensure they are independently coherent.

---

Provide in the following form:

Context: The entire context

Question: Regarding the entire context

Context 1: Sub-context 1

Context 2: Sub-context 2

---

Context: Trying to rebound from their divisional home loss to the Buccaneers, the Panthers flew to the Louisiana Superdome for a Week 5 divisional duel with the winless New Orleans Saints. With QB Jake Delhomme out and done for the year with a right elbow injury, QB David Carr was given the start. In the first quarter, Carolina took the early lead with kicker John Kasay getting a 23-yard field goal. The Saints responded with kicker Olindo Mare getting a 25-yard field goal. In the second quarter, the Panthers went back into the lead with Kasay nailing a 35-yard field goal. New Orleans would respond with Mare kicking a 28-yard field goal. In the third quarter, Carolina trailed as Saints FB Mike Karney got a 2-yard TD run for the only score of the period. In the fourth quarter, the Panthers tied the game with Carr completing a 17-yard TD pass to WR Steve Smith. Afterwards, Carolina sealed the win in the final seconds with Kasay nailing a 52-yard field goal as time ran out.

Question: How did the Carolina Panthers secure their victory against the New Orleans Saints in their Week 5 divisional duel?

Context 1: Trying to rebound from their divisional home loss to the Buccaneers, the Panthers flew to the Louisiana Superdome for a Week 5 divisional duel with the winless New Orleans Saints. With QB Jake Delhomme out for the year with a right elbow injury, QB David Carr was given the start. In the first quarter, Carolina took the early lead with kicker John Kasay getting a 23-yard field goal. The Saints responded with kicker Olindo Mare getting a 25-yard field goal. In the second quarter, the Panthers went back into the lead with Kasay nailing a 35-yard field goal, followed by New Orleans’ response with Mare kicking a 28-yard field goal.

Context 2: As the game progressed into the third quarter, the Panthers found themselves trailing after Saints FB Mike Karney got a 2-yard TD run, marking the only score of the period. However, in the fourth quarter, the Panthers managed to tie the game thanks to QB David Carr completing a 17-yard TD pass to WR Steve Smith. The climax of the match came in the final seconds with John Kasay nailing a 52-yard field goal as time ran out, securing a dramatic victory for Carolina against the New Orleans Saints.

---

Context: As a cell grows, its volume increases more quickly than its surface area. If a cell was to get very large, the small surface area would not allow enough nutrients to enter the cell quickly enough for the cell’s needs. However, large cells have a way of dealing with some size challenges. Big cells, such as some white blood cells, often grow more nuclei so that they can supply enough proteins and RNA for the cell’s requirements. Large, metabolically active cells often have lots of cell protrusions, resulting in many folds throughout the membrane. These folds increase the surface area available for transport of materials into or out of the cell. Such cell types are found lining your small intestine, where they absorb nutrients from your food through protrusions called microvilli.

Question: How do large cells adapt to the challenge of having a volume that increases more quickly than their surface area to meet their metabolic needs?

Context 1: As a cell grows, its volume increases more quickly than its surface area. If a cell was to get very large, the small surface area would not allow enough nutrients to enter the cell quickly enough for the cell's needs.

Context 2: Large cells have a way of dealing with their size challenges. Big cells, such as some white blood cells, often grow more nuclei so that they can supply enough proteins and RNA for the cell's requirements. Large, metabolically active cells often have lots of cell protrusions, resulting in many folds throughout the membrane. These folds increase the surface area available for transport of materials into or out of the cell. Such cell types are found lining your small intestine, where they absorb nutrients from your food through protrusions called microvilli.

---

Context: Philip Arnold Heseltine is best known as a composer of songs and other vocal music; he also achieved notoriety in his lifetime through his unconventional and often scandalous lifestyle.

Question: Why is Philip Arnold Heseltine's reputation mixed?

Context 1: Philip Arnold Heseltine is best known as a composer of songs and other vocal music.

Context 2: Philip Arnold Heseltine also achieved notoriety in his lifetime through his unconventional and often scandalous lifestyle.

---

Context: \textcolor{blue}{\{Context\}}

Question: 

\end{tcolorbox}

In the prompt template, the \{Context\} in blue color will be replaced with the given context to derive query and split during usage. After the LLM provides a response, we employ regular expressions to parse out the fields for Question, Context 1, and Context 2. Should the parsing fail, we will attempt the process up to three more times; otherwise, the context will be discarded. However, in our experiment, we find that almost all contexts can be successfully parsed in the first response. To observe this template in a more specific way, we have provided a detailed case demonstration in Appendix~\ref{appx:case-demo}.

For Chinese version: 

\definecolor{darkgreen}{rgb}{0,0.6,0}

\begin{tcolorbox}[breakable,colback=white!95!gray,colframe=gray!50!black,rounded corners,label={CST-template-Chinese}, title={Prompt Template for Context-Split-Tree with Instruction and Few-Shot in Chinese}]

\begin{CJK*}{UTF8}{gbsn}

给定整个段落Context，生成一个关于整个段落的用户可能关心的问题Question，问题答案要出自段落，然后将整个段落按语义划分为两个子段落，两子段落内进行一些必要的微调使得每一段相互独立。

---

以下面的形式给出：

Context: 整个段落

Question: 关于整个段落的问题

Context 1: 子段落1

Context 2: 子段落2

---

Context: 2020年11月11日，丁真去舅舅家吃饭时，偶然遇到了摄影师胡波。胡波本来想拍丁真的弟弟尼玛，但是没有遇到，所以胡波临时决定改拍丁真。然后胡波于当日将录制丁真的7秒短视频上传至抖音平台，丁真的短视频随即高热度传播，视频播放量过千万，点赞数百万。丁真珍珠随后在新浪微博上成为热门话题，相关话题阅读量达到五十亿。

Question: 丁真是怎么火起来的？

Context 1: 2020年11月11日，丁真去舅舅家吃饭时，偶然遇到了摄影师胡波。胡波本来想拍丁真的弟弟尼玛，但是没有遇到，所以胡波临时决定改拍丁真。

Context 2: 摄影师胡波将录制丁真的7秒短视频上传至抖音平台，丁真的短视频随即高热度传播，视频播放量过千万，点赞数百万。丁真珍珠随后在新浪微博上成为热门话题，相关话题阅读量达到五十亿。

---

Context: 当今世界，国际竞争越来越多地表现为知识产权的竞争。企业遭遇的海外知识产权纠纷日益激烈。我国已连续10年成为美国337调查的最大目标国，仅2012年就遭受337调查13起，数十家企业涉案。大力发展知识产权服务业，可增强市场主体的创新能力，促进品牌全球化，优化技术贸易结构，为企业实施“走出去”战略保驾护航。同时，大力发展版权等知识产权服务，有利于保护优秀创意成果，提升版权产业对国民经济的贡献率，促进文化产业和创意经济发展。而且，大力发展农产品地理标志和农作物种业等知识产权服务，可优化现代农业和林业的产业布局，拓宽农民增收渠道，促进城乡经济社会发展一体化。

Question: 发展知识产权服务业到底有啥用？

Context 1: 当今世界，国际竞争越来越多地表现为知识产权的竞争。企业遭遇的海外知识产权纠纷日益激烈。我国已连续10年成为美国337调查的最大目标国，仅2012年就遭受337调查13起，数十家企业涉案。大力发展知识产权服务业，可增强市场主体的创新能力，促进品牌全球化，优化技术贸易结构，为企业实施“走出去”战略保驾护航。

Context 2: 大力发展版权等知识产权服务，有利于保护优秀创意成果，提升版权产业对国民经济的贡献率，促进文化产业和创意经济发展。而且，大力发展农产品地理标志和农作物种业等知识产权服务，可优化现代农业和林业的产业布局，拓宽农民增收渠道，促进城乡经济社会发展一体化。

---

Context: 当前，高校所广泛采用的增量法对高校预算所具有的严肃性、权威性产生着一定的制约作用，因此高校应当以避免对预算进行频繁变更为出发点，采用零基预算、绩效预算等方式，做好中长期预算规划以及预算考评工作；另一方面高校有必要针对财务构建完善的激励机制与约束机制。在对预算执行情况开展绩效考评的基础上，高校不仅有必要将预算考评结果作为制定下一阶段预算规划的重要依据，而且需要构建起奖励制度与问责制度，通过鼓励良好的预算执行行为、追求预算执行不力现象当中的责任，对高校各个单位的预算执行行为进行激励与约束，从而确保财务预算的执行能够始终处于合理的范围之内。

Question: 现在高校是怎么保证财务预算处于合理范围内的？

Context 1: 当前，高校所广泛采用的增量法对高校预算所具有的严肃性、权威性产生着一定的制约作用，因此高校应当以避免对预算进行频繁变更为出发点，采用零基预算、绩效预算等方式，做好中长期预算规划以及预算考评工作。

Context 2: 现在高校有必要针对财务构建完善的激励机制与约束机制。在对预算执行情况开展绩效考评的基础上，高校不仅有必要将预算考评结果作为制定下一阶段预算规划的重要依据，而且需要构建起奖励制度与问责制度，通过鼓励良好的预算执行行为、追求预算执行不力现象当中的责任，对高校各个单位的预算执行行为进行激励与约束，从而确保财务预算的执行能够始终处于合理的范围之内。

---

Context: \textcolor{blue}{\{Context\}}

Question: 

\end{CJK*}
\end{tcolorbox}

Note that the three few-shot examples in the prompt templates are not entirely fixed but can be adapted to different corpus by selecting suitable examples to stimulate the model to generate a question distribution that is close to the distribution of real user questions about that domain corpus, which can improve the quality of the synthetic data and the effectiveness of the final fine-tuned model.

\subsection{Training Scorer to Rank Queries and Filtering}
\label{appx:detail-step2}


\subsubsection{Training Scorer via Contrastive Learning}
\label{appx:step2-training}

In our approach to enhancing data quality and diversity, we focus on the innovative construction of positive and negative samples for training our scorer. By employing contrastive learning, we train the scorer that efficiently evaluates the degree of adherence to instruct prompts and few-shot examples, surpassing previous methods that rely on heuristic algorithms or direct scoring, which are prone to positional bias and instability. 

Positive samples are straightforwardly generated by employing the LLM to create context-query pairs that adhere to well-designed instruct prompts and few-shot examples shown in Appendix~\ref{appx:CST-prompt-template}, which are also the queries generated through Step~1. These serve as exemplars reflecting the desired output. On the other hand, the creation of negative samples involves intentional manipulation of instruct prompts or few-shot examples, or both. More specifically, we manipulate the instruction by simplifying the instruct prompt to \textit{“Given a context, generate a question and split context into two sub-contexts”}. To manipulate few-shot examples, we degrading it to one-shot by retaining only one example. We also combine both approaches to simultaneously manipulate the instruction and few-shot examples to generate more negative samples. These manipulations aim to deviate from the optimal query generation, thus producing examples that diverge from the model's training objective. For each type, we randomly select $500$ positive samples generated from Step~1 to construct negative samples, forming $1,500$ positive-negative sample pairs for scorer training in total. 

We construct the scorer's structure as outlined in Section~\ref{sec:step2}, initializing it with parameters from our base LLM to serve as the training warm-up. The scorer's parameter set is a duplicated version, ensuring that modifications do not impact the original base LLM. For efficient training, we employ QLoRA with 4-bit quantization and the ranks of $32$, significantly reducing GPU memory requirements and speeding up the training process. A more detailed breakdown of the hyperparameter configuration is provided in Table~\ref{tab:hyper-training}.

\subsubsection{Collaborating Scorer with CST to Filter Queries}
\label{appx:step2-filter}

The scorer is designed to work in cooperation with CST to enhance the quality and diversity of the generated questions while also meeting the quantitative requirements. For a given context, we first use CST to generate a series of potential questions. Then, each question is scored by the scorer, with the scores used to rank them from highest to lowest. We sequentially add questions to a candidate set, but only if the current question's ROUGE-L F1 similarity to any question already in the set is less than~$0.7$. This process continues until the number of questions in the candidate set reaches the desired quantity~$N$.

If the initial round of CST and scoring does not yield the required number of questions, we initiate another round of CST to expand on the initial questions, followed by repeating the scoring and selection process until the target quantity is achieved. The scorer's role in scoring and filtering is like to effectively condense the output from CST, ensuring that even with a smaller set of questions, both high quality and diversity are maintained. The pseudocode of filtering process is shown in Algorithm~\ref{algo:filtering}.

\begin{center}
\begin{minipage}{0.8\textwidth}
\begin{algorithm}[H]
\caption{Scorer collaborates with CST to filter queries}\label{algo:filtering}
\textbf{Input:} A Context $C$, Required number of maintained queries $N$ \\
\textbf{Output:} Query dataset $Data$ comprises exactly $N$ queries with high quality and diversity
\begin{algorithmic}[1]

\Function{Filter}{$Q_{All}$}
    \State Initialize $Q_{Cand} \gets \text{empty list}$
    \State Sort $Q_{All}$ by score descending
    \For{each $(q, s) \in Q_{All}$}
        \If{All ROUGE-L[F1] with $Q_{Cand} < 0.7$}
            \State Append $q$ to $Q_{Cand}$ \Comment{Append if diversity reach the threshold}
            \If{$len(Q_{Cand}) = N$}
                \Return{$Q_{Cand}$} \Comment{Enough quantity}
            \EndIf
        \EndIf
    \EndFor
    \State \Return{$Q_{Cand}$}
\EndFunction
\State
\State Initialize $Data \gets \text{empty list}$
\State Initialize $Q_{All} \gets \text{empty list}$ \Comment{Store all query-score pairs}
\While{$len(Data) < N$} \Comment{Iterate until enough queries obtained}
    \State Initialize $Q_{New} \gets \text{empty list}$
    \State \Call{ContextSplitTree}{$C, Q_{New}$} \Comment{Call CST to get new queries}
    \For{each $q \in Q_{New}$}
        \State Apeend $(q,Sc(C,q))$ to $Q_{All}$ \Comment{Score each new query}
    \EndFor
    \State Set $Data \gets$ \Call{Filter}{$Q_{All}$}
\EndWhile
\State \Return{$Data$}

\end{algorithmic}
\end{algorithm}
\end{minipage}
\end{center}

\subsection{Obtaining High-Fidelity Responses}
\label{appx:step3}

Our design is motivated by the significant influence principles have on guiding LLMs, aiming to achieve high-fidelity responses through a principle-driven self-alignment step. These principles are anticipated to enhance the LLM's ability to produce high-fidelity, realistic, and helpful answers, and the specific principles vary depending on the task and remain exploratory. They may also include rules for directing the LLM to generate responses in a particular tone. This could be particularly valuable when creating SFT data for a custom LLM assistant or for role-playing. Furthermore, the existence of principles serves as a method for aligning with human preferences, offering a viable alternative to the cumbersome process of reinforcement learning from human feedback (RLHF)~\cite{ouyang2022training}.

Different from previous approaches, we innovate to integrate a self-improving pipeline to further increase fidelity. Instead of manually selecting a few-shot examples from annotated examples, we divide the annotated examples into training and testing sets. We then conduct a random search that iteratively selects a subset from the training set and allows the LLM to self-evaluate the output scores in the test set. This process is iterated $16$ times by default, and the subset that achieves the highest scores in the test set is used as the few-shot ICL examples. We implement this pipeline through the DSPy~\cite{khattab2023dspy} framework, significantly reducing coding effort. This self-improving process works well with principle-driven self-alignment, as it aids in identifying the optimal ICL examples that guide the LLM to generate helpful, realistic, and reliable answers in line with alignment principles, markedly enhancing the quality and fidelity of the responses and, consequently, the responses by the fine-tuned models.

Finally, we prune all contexts, principles, and ICL examples to retain only the query-response pairs for supervised fine-tuning of the LLM. While several studies~\cite{zhang2024recost,yu2023wavecoder,xia2024less} try to further execute filtering on generated answers, we leave it as a future work as it is not such crucial for our method. Actually, simply generating additional iterations on the same query and retaining self-consistent~\cite{wang2022self} responses may further improve some degrees of reasoning accuracy for short-form responses. However, this might not be a good deal when also taking the computing costs into consideration since letting LLMs improve and correct their own responses is not an easy thing~\cite{huang2023large}. We believe that since each question we obtain in CST precisely matches the granularity of its context, it will be easy for LLM to provide accurate and pertinent answers to the questions.

\section{Implementation Details}
\label{appx:implementation-details}

All experiments are implemented on a single node with eight Nvidia A100 80G GPUs and $160$ Intel Xeon Gold 6248 CPUs. To speed up the generation of LLM calls, we use the vLLM~\cite{kwon2023efficient} inference engine for acceleration, and make concurrent requests with a concurrency of $8$ threads. In order to reduce GPU memory usage and accelerate training speed, we use the DeepSpeed~\cite{rasley2020deepspeed} distributed training framework accelerating with ZeRO-2~\cite{rajbhandari2020zero}, where the AdamW~\cite{loshchilov2017decoupled} optimizer is applied for gradient descent. We employ QLoRA with 4-bit quantization and the ranks of $32$, which has been demonstrated to be able to achieve satisfactory results in previous works~\cite{jiang2024improving,zhang2023machine}. We train for 4 epochs in total, for it has been shown as the maximum number of iterations that negligible affect the training loss~\cite{muennighoff2024scaling}. On the \datasetname{} dataset, our \methodname{} generates about $120$K pieces of data in $184$ A100 GPU hours and required another $272$ A100 GPU hours for supervised fine-tuning. On four benchmarks, the generation throughput is about $340$ pairs per A100 GPU hour and the total running times vary from $120$ to $448$ A100 GPU hours depending on corpus size. More detailed settings on hyperparameters can be found in Appendix~\ref{appx:hyperparameters}.

\subsection{Assets Use}
\label{appx:assets}

All the pre-trained LLM and open-source datasets we use in experiments can be respectively found on Huggingface transformers~\cite{wolf2019huggingface} and datasets~\cite{lhoest2021datasets}, and we have checked that they are available for research purposes and have been properly cited and correctly adhered to open-source licenses. We list the public links of the used LLMs in Table~\ref{tab:link-llm} and the used datasets in Table~\ref{tab:link-dataset}. We will also make our \datasetname{} dataset open-sourced at \githublink{} to boost the academy.

\begin{table}[h]
    \caption{Public links to the used LLMs.}
    \label{tab:link-llm}
    \centering
    \begin{tabular}{l|l}\toprule
        \textbf{LLM} & \textbf{Link} \\ \midrule
        Qwen1.5-c$_{\textsc{32b}}$~\cite{bai2023qwen} & \url{https://huggingface.co/Qwen/Qwen1.5-32B-Chat} \\
        Llama3-c$_{\textsc{70b}}$~\cite{llama3modelcard} & \url{https://huggingface.co/meta-llama/Meta-Llama-3-70B-Instruct} \\
         \bottomrule
    \end{tabular}
\end{table}

\begin{table}[H]
    \caption{Public links to the used datasets.}
    \label{tab:link-dataset}
    \centering
    \begin{tabular}{l|l}\toprule
        \textbf{Dataset} & \textbf{Link} \\ \midrule
        SQuAD1.1~\cite{rajpurkar2016squad} & \url{https://huggingface.co/datasets/rajpurkar/squad} \\
        TriviaQA~\cite{joshi2017triviaqa} & \url{https://huggingface.co/datasets/mandarjoshi/trivia_qa} \\
        DROP~\cite{dua2019drop} &
        \url{https://huggingface.co/datasets/ucinlp/drop} \\
        WebGLM-QA~\cite{liu2023webglm} & \url{https://huggingface.co/datasets/THUDM/webglm-qa} \\
         \bottomrule
    \end{tabular}
\end{table}

\subsection{Hyperparameters}
\label{appx:hyperparameters}

We present the generation hyperparameters in Table~\ref{tab:generation-hyper} and the fine-tuning configurations in Table~\ref{tab:hyper-training}.

\begin{table}[h]
    \caption{Generation Hyperparameters}
    \label{tab:generation-hyper}
    \centering
    \begin{tabular}{l|l}\toprule
        \textbf{Parameter} & \textbf{Value} \\ \midrule
        Max instruction length & 4096 \\
        Max new tokens & 4096 \\
        Top-k & 50 \\
        Top-p & 1.0 \\
        Temperature (for query) & 0.85 \\
        Temperature (for response) & 0.2 \\
         \bottomrule
    \end{tabular}
\end{table}

\begin{table}[h]
    \caption{Training Configurations}
    \label{tab:hyper-training}
    \centering
    \begin{tabular}{l|l}\toprule
        \textbf{Parameter} & \textbf{Value} \\ \midrule
        Epoch & 4 \\
        Learning rate & 5e-5 \\
        Mini batch size & 4 \\
        Warmup steps  & 50 \\
        Weight decay & 0.01 \\
        Compute dtype & bfloat16 \\
        Quantization dtype & nf4 \\
        Lora rank & 32 \\
        Lora alpha & 32 \\
        Lora dropout & 0.05 \\
        Lora bias & none \\
         \bottomrule
    \end{tabular}
\end{table}

\section{Human Evaluation Guidance}
\label{appx:guidance}

To establish a robust assessment framework for both generated queries and model outputs, we have devised an extensive human evaluation guideline shown in Table~\ref{tab:guidance}. Each score will also be accompanied by several corresponding examples, ensuring a consistent and objective evaluation process. This guideline emphasizes key metrics, including realism, diversity, relevance, accuracy, and satisfaction. By following this guide, evaluators can thoroughly assess the effectiveness of our method, guaranteeing the generation of high-quality, multi-granularity queries and responses. Our approach strives for comprehensive evaluations, aided by detailed scoring rubrics and examples to enable balanced decision-making.

\newgeometry{top=1.8cm,bottom=2.5cm}
\begin{table}[ht]
\centering
\caption{The guidance for human evaluation.}
\label{tab:guidance}

\begin{tabular}{>{\raggedright\arraybackslash}p{0.1\linewidth} >{\raggedright\arraybackslash}p{0.8\linewidth}}
\toprule
\textbf{Score} & \textbf{Realism} \\
\midrule
5 & The query is indistinguishable from those a human might ask. It is natural, authentic, and precisely the type of question a curious user would pose.  \\
\rowcolor{gray!25}
4 & The query closely resembles real user inquiries, with minor differences. It maintains a high level of realism and naturalness. \\
3 & The query shows moderate realism, differing somewhat from typical user questions. It still appears natural and understanding. \\
\rowcolor{gray!25}
2 & The query has noticeable deviations from real user questions, affecting its realism. It shows signs of artificiality but remains understandable. \\
1 & The query is clearly artificial, lacking realism and naturalness. It differs significantly from how a real user would ask. \\
\bottomrule
\end{tabular}
\begin{tabular}{>{\raggedright\arraybackslash}p{0.1\linewidth} >{\raggedright\arraybackslash}p{0.8\linewidth}}
\toprule
\textbf{Score} & \textbf{Diversity} \\
\midrule
5 & The queries exhibit exceptional diversity, covering a wide range of topics and varying greatly in their nature and specificity. \\
\rowcolor{gray!25}
4 & The queries show good diversity, exploring multiple topics and presenting different types of questions. They maintain a solid variety, even if not exhaustive. \\
3 & The queries present moderate diversity, touching upon several topics but with some repetitiveness or predictability in their nature. \\
\rowcolor{gray!25}
2 & The queries show limited diversity, often sticking to a narrow range of topics or lacking variety in their structure and content. \\
1 & The queries lack diversity, being highly repetitive, monotonous, and showing minimal to no variation in topics or approach. \\
\bottomrule
\end{tabular}
\begin{tabular}{>{\raggedright\arraybackslash}p{0.1\linewidth} >{\raggedright\arraybackslash}p{0.8\linewidth}}
\toprule
\textbf{Score} & \textbf{Relevance} \\
\midrule
5 & The response is highly relevant, precisely addressing the query’s intent and providing contextually appropriate information. \\
\rowcolor{gray!25}
4 & The response is mostly relevant, with minor deviations that do not significantly affect its overall alignment with the query. \\
3 & The response shows moderate relevance, partially addressing the query but with some noticeable gaps or misalignments. \\
\rowcolor{gray!25}
2 & The response has limited relevance, straying significantly from the core of the query or providing only partially related information. \\
1 & The response is irrelevant, failing to address the query’s intent or providing information that is completely off-topic. \\
\bottomrule
\end{tabular}

\begin{tabular}{>{\raggedright\arraybackslash}p{0.1\linewidth} >{\raggedright\arraybackslash}p{0.8\linewidth}}

\toprule
\textbf{Score} & \textbf{Accuracy} \\
\midrule
5 & The response is completely accurate, with no factual errors or hallucinations. All information provided is verifiable and aligns with external sources. \\
\rowcolor{gray!25}
4 & The response contains minor inaccuracies or minor hallucinations, but the overall information conveyed is mostly correct and reliable. \\
3 & The response shows moderate accuracy, with some noticeable factual errors or hallucinations that don’t significantly alter the main message. \\
\rowcolor{gray!25}
2 & The response has significant inaccuracies or hallucinations, affecting the overall reliability and correctness of the information provided. \\
1 & The response is highly inaccurate, containing multiple factual errors or severe hallucinations that render the information untrustworthy. \\
\bottomrule
\end{tabular}

\begin{tabular}{>{\raggedright\arraybackslash}p{0.1\linewidth} >{\raggedright\arraybackslash}p{0.8\linewidth}}
\toprule
\textbf{Score} & \textbf{Satisfaction}\\
\midrule
5 & The evaluator is highly satisfied with the responses. They fully meet expectations, leaving no room for improvement.\\
\rowcolor{gray!25}
4 & The evaluator is mostly satisfied. The responses are generally good, with only minor shortcomings or areas for improvement.\\
3 & The evaluator feels moderately satisfied. The responses have notable strengths but also some weaknesses that need addressing.\\
\rowcolor{gray!25}
2 & The evaluator is somewhat dissatisfied. The responses show significant room for improvement and may not fully meet expectations.\\
1 & The evaluator is highly dissatisfied. The responses fail to meet expectations on multiple levels, requiring substantial improvement.\\
\bottomrule
\end{tabular}

\end{table}

\newgeometry{
    textheight=9in,
    textwidth=5.5in,
    top=1in,
    headheight=12pt,
    headsep=25pt,
    footskip=30pt
}

\section{Case Demonstration}
\label{appx:case-demo}

In this section, we select a specific context as a case demonstration to inspect the entire process around it, from generating questions by constructing Context-Split-Tree, to scoring queries and filtering them, and finally to generating answers by attaching self-alignment principles and optimal few-shot examples. This demonstration is designed to provide a clearer and more intuitive understanding of our algorithm's workflow.

Initially, we input the starting context as the root node of the whole Context-Split-Tree. With the assistance of the LLM, we derive an overall question from this context and simultaneously split it into two child contexts. We have omitted the specific instruction in the text boxes below, as it has already been presented in detail in Appendix~\ref{appx:CST-prompt-template}. The Context in boxes is the input while the Question, Context 1, and Context 2 are parsed outputs:

\begin{tcolorbox}[breakable,colback=white!95!gray,colframe=gray!50!black,rounded corners,label={template-chinese-demo}, title={Node 1 $\rightarrow$ Node 2, Node 3}]
\textbf{Context: }The profits of the contemporary global value chains (GVC) form a V-shape, also known as the “smile curve”. At one end of this curve are research and development (R\&D) and design, and at the other end are services and marketing, with processing and production situated in the middle. Typically, the profit margin for industries at both ends ranges from 20\% to 25\%, whereas the profit margin for the production processes in the middle stands at merely 5\%.

\textbf{Question: }Why do entrepreneurs worldwide strive to move up the value chain?

\textbf{Context 1: }The profits of the contemporary global value chains (GVC) form a V-shape, also known as the “smile curve”. At one end of this curve are research and development (R\&D) and design, and at the other end are services and marketing, with processing and production situated in the middle.

\textbf{Context 2: }Typically, the profit margin for industries at both ends of the global value chains ranges from 20\% to 25\%, whereas the profit margin for the production processes in the middle stands at merely 5\%.

\end{tcolorbox}

The child context undergoes recursive splitting and derives questions with corresponding granularity: 

\begin{tcolorbox}[breakable,colback=white!95!gray,colframe=gray!50!black,rounded corners,label={template-chinese-demo}, title={Node 2 $\rightarrow$ Node 4, Node 5}]

\textbf{Context: }The profits of the contemporary global value chains (GVC) form a V-shape, also known as the “smile curve”. At one end of this curve are research and development (R\&D) and design, and at the other end are services and marketing, with processing and production situated in the middle.

\textbf{Question: }What are the key components of the contemporary global value chains?

\textbf{Context 1: }The profits of the contemporary global value chains (GVC) form a V-shape, also known as the “smile curve”.

\textbf{Context 2: }At one end of the smile curve are research and development (R\&D) and design, and at the other end are services and marketing, with processing and production situated in the middle.

\end{tcolorbox}

We use depth-first-search (DFS) so this is the time for Node 4. Actually, using breadth-first-search (BFS) will ultimately achieve the same effect. Node 4 derives a question but terminates afterward due to its unsuccessful attempt to split into two viable child contexts:

\begin{tcolorbox}[breakable,colback=white!95!gray,colframe=gray!50!black,rounded corners,label={template-chinese-demo}, title={Node 4 Ends}]

\textbf{Context: }The profits of the contemporary global value chains (GVC) form a V-shape, also known as the “smile curve”.

\textbf{Question: }What does the global value curve look like?

\textbf{Context 1: }The profits of the contemporary global value chains (GVC) form a V-shape, also known as the “smile curve”.

\textbf{Context 2: }

\end{tcolorbox}

The child Context 1 split from Node 5 will not further derive questions or split because it falls below the minimum length threshold. However, one can adjust this threshold to a lower value to derive more detailed questions:

\begin{tcolorbox}[breakable,colback=white!95!gray,colframe=gray!50!black,rounded corners,label={template-chinese-demo}, title={Node 5 $\rightarrow$ Node 6}]

\textbf{Context: }At one end of the smile curve are research and development (R\&D) and design, and at the other end are services and marketing, with processing and production situated in the middle.

\textbf{Question: }What is the structure of the smile curve?

\textbf{Context 1: }At one end of the smile curve are research and development (R\&D) and design.

\textbf{Context 2: }The other end of the smile curve are services and marketing, with processing and production situated in the middle.

\end{tcolorbox}

Node 6 derives a question and terminates because both of its child contexts are too short:

\begin{tcolorbox}[breakable,colback=white!95!gray,colframe=gray!50!black,rounded corners,label={template-chinese-demo}, title={Node 6 Ends}]

\textbf{Context: }The other end of the smile curve are services and marketing, with processing and production situated in the middle.

\textbf{Question: }What lies in the middle of the smile curve?

\textbf{Context 1: }The other end of the smile curve are services and marketing.

\textbf{Context 2: }The processing and production are situated in the middle.

\end{tcolorbox}

After Node 2 has completed its recursion, it is now Node 3's turn to proceed:

\begin{tcolorbox}[breakable,colback=white!95!gray,colframe=gray!50!black,rounded corners,label={template-chinese-demo}, title={Node 3 $\rightarrow$ Node 7, Node 8}]
\textbf{Context: }Typically, the profit margin for industries at both ends of the global value chains ranges from 20\% to 25\%, whereas the profit margin for the production processes in the middle stands at merely 5\%.

\textbf{Question: }Which type of industry has the lowest profit margin?

\textbf{Context 1: }Typically, the profit margin for industries at both ends of the global value chains ranges from 20\% to 25\%.

\textbf{Context 2: }Whereas the profit margin for the production processes in the middle stands at merely 5\%.
\end{tcolorbox}

Node 7 and Node 8 terminate after deriving one detailed question each, as they have reached the minimum granularity and cannot split properly:

\begin{tcolorbox}[breakable,colback=white!95!gray,colframe=gray!50!black,rounded corners,label={template-chinese-demo}, title={Node 7 Ends}]
\textbf{Context: }Typically, the profit margin for industries at both ends of the global value chains ranges from 20\% to 25\%.

\textbf{Question: }How high can the profit margin go for industries at two ends of the global value chains?

\textbf{Context 1: }Typically, the profit margin for industries at both ends of the global value chains ranges from 20\% to 25\%.

\textbf{Context 2: }
\end{tcolorbox}

\begin{tcolorbox}[breakable,colback=white!95!gray,colframe=gray!50!black,rounded corners,label={template-chinese-demo}, title={Node 8 Ends}]
\textbf{Context: }Whereas the profit margin for the production processes in the middle stands at merely 5\%.

\textbf{Question: }What is the profit margin for the production processes?

\textbf{Context 1: }Whereas the profit margin for the production processes in the middle stands at merely 5\%.

\textbf{Context 2: }
\end{tcolorbox}

\begin{figure}[h]
    \centering
    \includegraphics[width=0.7\linewidth]{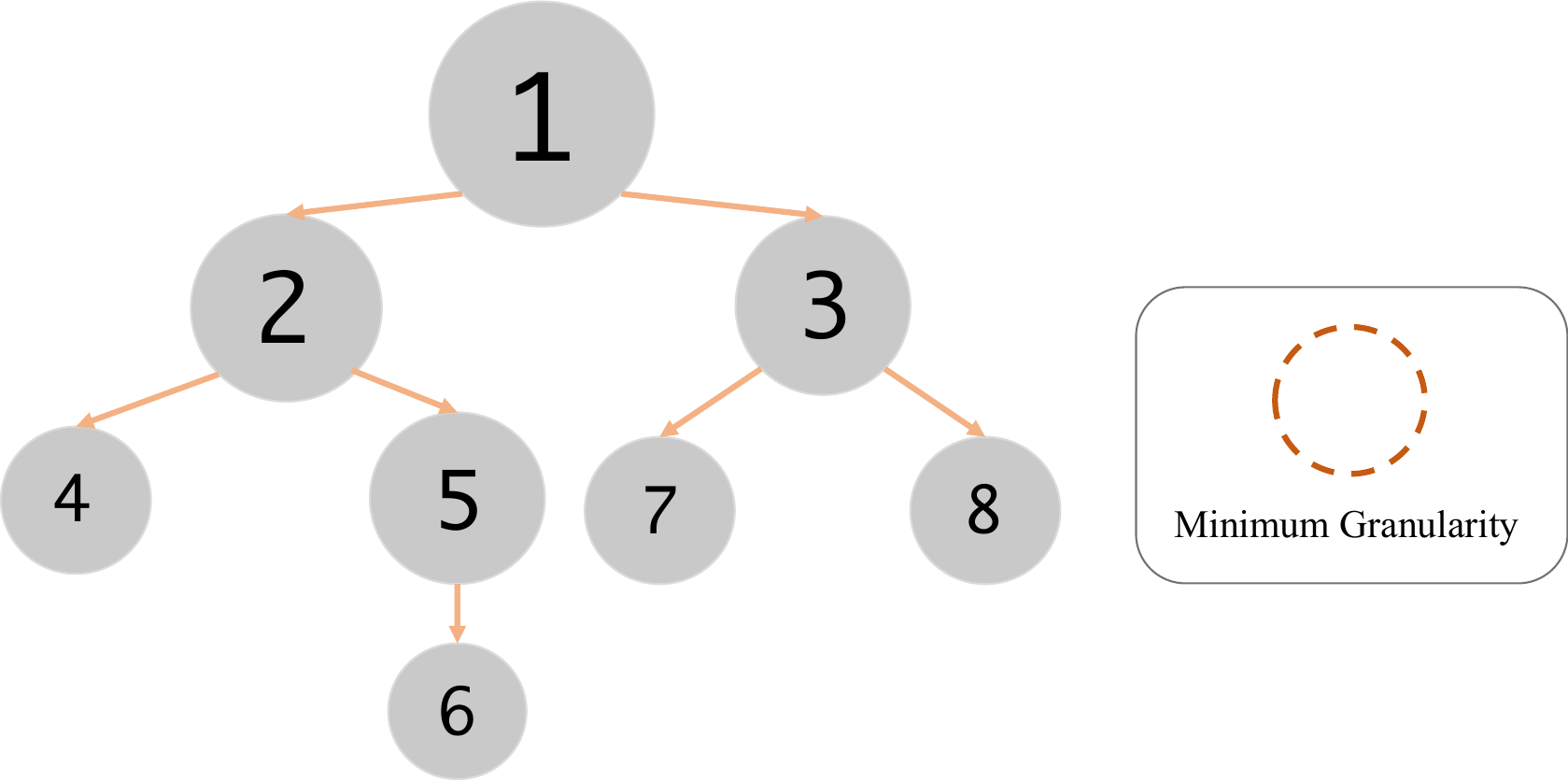}
    \caption{The schematic of the constructed CST in this case. Each node contains a context and a corresponding question, with the node size indicating different levels of granularity.}
    \label{fig:CST-demo}
\end{figure}

Then, the recursion comes to an end, and this entire process ultimately results in the formation of the Context-Split-Tree depicted in Figure~\ref{fig:CST-demo}. Each node within this tree contains a context and a question that align with the corresponding granularity.

We collect context question pairs from all nodes into a list and employ the trained scorer to evaluate each item. The items are then sorted based on their scores, from highest to lowest. Next, we sequentially examine each item and choose those queries whose ROUGE-L scores with all previously selected queries are below 0.7. Due to the low ROUGE-L scores among query pairs in that case, the resulting selected set shown in Table~\ref{tab:queries} primarily comprises the preceding few items. However, if we aim to derive a greater number of questions from the context, such as 10 questions, an additional round of CST becomes necessary. Following this, all queries generated across both CST sessions undergo a collective ranking and filtering process. During this step, the ROUGE-L metric proves useful for eliminating queries that have lower scores and are similar to previously selected ones.


At this point, in addition to our results, we also provide the generated query list using the Context-Instruct method for a direct comparison. The Context-Instruct method produces question-confidence-answer triplets, where the confidence level can be either high or low and is used for filtering purposes. The results are presented in Table~\ref{tab:queries-context-instruct}, with the responses omitted to conserve space. Putting Table~\ref{tab:queries} and Table~\ref{tab:queries-context-instruct} together, we can perceive that the queries generated by Context-Instruct exhibit a noticeable lack of diversity and granularity compared to our queries. This discrepancy provides an intuitive explanation for our method's superior performance in generating multi-granularity queries and ultimately producing better results than other methods.

\begin{table}[h]
\centering
\begin{tabular}{p{0.8\textwidth}cc}
\toprule
\textbf{query} & \textbf{score} & \textbf{select} \\
\midrule
Why do entrepreneurs worldwide strive to move up the value chain? & 0.95 & \ding{51} \\
What are the key components of the contemporary global value chains? & 0.91 & \ding{51} \\
How high can the profit margin go for industries at two ends of the global value chains? & 0.88 & \ding{51} \\
What does the global value curve look like? & 0.83 & \ding{51} \\
Which type of industry has the lowest profit margin? & 0.74 & \ding{55} \\
What is the structure of the smile curve? & 0.67 & \ding{55} \\
What is the profit margin for the production processes? & 0.64 & \ding{55} \\
What lies in the middle of the smile curve? & 0.59 & \ding{55} \\
\bottomrule
\end{tabular}
\caption{The results of ranking and filtering the queries.}
\label{tab:queries}
\end{table}

\begin{table}[h]
\centering
\begin{tabular}{p{0.75\textwidth}cc}
\toprule
\textbf{query} & \textbf{confidence} & \textbf{select} \\
\midrule
What is the shape of the profits of global value chains? & High & \ding{51} \\
What is the “smile curve” also known as? & High & \ding{51} \\
What is the primary function of the industries at one end of the “smile curve”? & High & \ding{51} \\
Which activities have a profit margin ranging from 20\% to 25\%? & High & \ding{51} \\
What is the typical profit margin for industries in the middle of the “smile curve”? & High & \ding{55} \\
What is situated at both ends of the “smile curve”? & Low & \ding{55} \\
Do companies typically earn higher profits from production processes than from R\&D? & Low & \ding{55} \\
Is the “smile curve” a linear representation of profits in global value chains? & Low & \ding{55} \\
\bottomrule
\end{tabular}
\caption{The generated questions using Context-Instruct.}
\label{tab:queries-context-instruct}
\end{table}

Returning to our workflow, after filtering the queries, we leverage the alignment principles and a select few-shot examples (detailed in Appendix 5) to assist our LLM in providing high-fidelity answers to these questions. Thanks to our Context-Split-Tree process providing the specific window of context required to answer the questions, obtaining these answers is relatively easy and results in more accurate and context-relevant responses. Once we have obtained high-fidelity answers, we remove the alignment principles, few-shot examples, and context, retaining only the query-response pairs to supervised fine-tune our LLM. The final obtained SFT data for this case are presented in the following text tables. 

\begin{tcolorbox}[breakable,colback=white!95!gray,colframe=gray!50!black,rounded corners,label={template-chinese-demo}, title={SFT Data 1}]
\textbf{Question: }Why do entrepreneurs worldwide strive to move up the value chain? 

\textbf{Answer: }Entrepreneurs worldwide strive to move up the value chain because the profit margins are significantly higher at the ends of the curve, ranging from 20\% to 25\%, compared to the middle, which has a profit margin of only 5\%. By moving up the value chain, entrepreneurs can increase their profit margins and gain a competitive advantage in the market.
\end{tcolorbox}

\begin{tcolorbox}[breakable,colback=white!95!gray,colframe=gray!50!black,rounded corners,label={template-chinese-demo}, title={SFT Data 2}]
\textbf{Question: }What are the key components of the contemporary global value chains? 

\textbf{Answer: }The key components of the contemporary global value chains are:

1. Research and Development (R\&D) and Design (at one end of the curve)

2. Processing and Production (at the other end of the curve)

3. Services and Marketing (in the middle of the curve)
\end{tcolorbox}

\begin{tcolorbox}[breakable,colback=white!95!gray,colframe=gray!50!black,rounded corners,label={template-chinese-demo}, title={SFT Data 3}]
\textbf{Question: }How high can the profit margin go for industries at two ends of the global value chains? 

\textbf{Answer: }The profit margin for industries at both ends of the global value chains can go up to 25\%.
\end{tcolorbox}

\begin{tcolorbox}[breakable,colback=white!95!gray,colframe=gray!50!black,rounded corners,label={template-chinese-demo}, title={SFT Data 4}]
\textbf{Question: }What does the global value curve look like? 

\textbf{Answer: }It looks like a V-shape, also known as the “smile curve”.
\end{tcolorbox}

\section{Additional Experiments}
\label{appx:addtional-experiments}

\subsection{Ablation Study}
\label{appx:ablation-study}

In this section, we detail the ablation experiments conducted to assess the indispensability and impact of the three essential steps in our proposed method. These steps are integral to our approach, designed to recursively derive queries, rank and filter them for quality and diversity, and finally, generate high-fidelity responses. Through these experiments, we aim to delineate the contribution of each step towards the overall effectiveness of our method. In this study, we develop the following four distinct variations of our method, with each one specifically tailored to concentrate on a fundamental step:

\begin{enumerate}[leftmargin=*]

    \item \textbf{$\methodname{}^{w/o}_{\text{CST1}}$} omits the use of the Context-Split-Tree for iteratively splitting and generating queries for given contexts. Instead, $\methodname{}^{w/o}_{\text{CST}}$ employs a technique where few-shot examples are used to iteratively derive queries from the extracted context until the desired number of queries is obtained (we set the desired number to be the same with all generated queries of \methodname{} without filtering). The purpose of this modification is to assess the efficacy of CST in deriving multi-granularity queries. Additionally, this variant facilitates an examination of how the exclusion of CST impacts the diversity of the generated queries and the overall performance of the final fine-tuned model.

    \item \textbf{$\methodname{}^{w/o}_{\text{CST2}}$} also omits the use of the Context-Split-Tree for iteratively splitting and generating queries for given contexts. Different from $\methodname{}^{w/o}_{\text{CST1}}$, $\methodname{}^{w/o}_{\text{CST2}}$ splits the contexts in a heuristic way that each time splits it in the middle (we will let the whole sentence in the middle in the first sub context to maintain semantic integrity) until reaching the minimum granularity. And then use all split contexts to iteratively derive queries until the quantity is enough. This variant is designed to further assess the efficacy of CST in deriving multi-granularity queries with the comparison with a heuristic context segmentation method.

    \item \textbf{$\methodname{}^{w/o}_{\text{filter}}$} eliminates the scoring and filtering process to evaluate its effects on the overall quality and diversity of the generated queries. If the number of queries generated in Step 1 exceeds the predetermined limit, we just proceed by randomly selecting a sufficient number of queries to meet the quota. This variant enables us to assess the effects of bypassing our established quality and diversity control mechanisms.

    \item \textbf{$\methodname{}^{w/o}_{\text{fidelity}}$} obtains the answers to the queries without adhering to self-alignment or employing the self-improving. Instead, $\methodname{}^{w/o}_{\text{fidelity}}$ utilizes fixed predetermined few-shot examples along with a straightforward prompt design devoid of guiding principles. This variant allows us to evaluate the efficacy of our response generation methodology in enhancing the overall quality and relevance of responses.
\end{enumerate}

We implement the four variants on TriviaQA (short-form) and WebGLM-QA (long-form) datasets and conduct a comparison with our \methodname{}. The results are shown in Table~\ref{tab:ablation}.

\begin{table*}[h]
\centering
\begin{tabular}{lrr}\toprule
&\multicolumn{1}{c}{Short-form (Acc)} & \multicolumn{1}{c}{Long-form (BS)}\\
Variant&TriviaQA&WebGLM-QA\\
\midrule

$\methodname{}^{w/o}_{\text{CST1}}$&
    $0.793 \scriptstyle \pm 0.003$&
    $0.912 \scriptstyle \pm 0.001$ \\
$\methodname{}^{w/o}_{\text{CST2}}$&
    $0.826 \scriptstyle \pm 0.003$&
    $0.910 \scriptstyle \pm 0.001$ \\
$\methodname{}^{w/o}_{\text{filter}}$&
    $0.828 \scriptstyle \pm 0.003$&
    $0.915 \scriptstyle \pm 0.001$ \\
$\methodname{}^{w/o}_{\text{fidelity}}$&
    $0.833 \scriptstyle \pm 0.004$&
    $0.907 \scriptstyle \pm 0.002$ \\
\methodname{}&
    $\mathbf{0.849} \scriptstyle \pm 0.003$&
    $\mathbf{0.924} \scriptstyle \pm 0.002$ \\
\bottomrule
\end{tabular}
\caption{The results of ablation study.}
\label{tab:ablation}
\end{table*}

Our analysis has led to three key insights. Firstly, when compared to our \methodname{}, all variants yield suboptimal outcomes. This highlights the critical nature of each step within our methodology, underscoring the fact that they are all crucial and collectively contribute to achieving superior performance.

Secondly, within the context of short-form datasets, it was observed that the variants that undergo modifications in the CST process perform the poorest. This finding suggests that the CST process plays a vital role in encompassing a comprehensive scope of granularity, thereby enabling the extraction of a broader spectrum of knowledge.

Thirdly, with regard to long-form datasets, the variant $\methodname{}^{w/o}_{\text{fidelity}}$ demonstrates the lowest level of performance. This outcome underlines the significance of self-alignment and self-enhancement mechanisms in generating responses of high quality and fidelity.

\subsection{GPT-4 Judge}
\label{appx:gpt-4-judge}

To conduct a more comprehensive evaluation, we utilized GPT-4~\cite{achiam2023gpt} (we use gpt-4-0125-preview) to judge as the referee on the baseline method and our approach. By using GPT-4 to compare the outputs of the model fine-tuned on our generated data with the baseline methods, we can better understand the advantages of our model and avoid biases stemming from manual preferences. We provide the query in the \datasetname{} test set as inputs and get the outputs of the model fine-tuned using our method and baseline method on the \datasetname{} dataset. To facilitate a nuanced evaluation, we categorized the queries into three levels of granularity: detail question (\textit{e.g.}, \textit{how deep does the abortion needle penetrate?}), concept question (\textit{e.g.}, \textit{what is the difference between emergencies and crises?}), and macro question (\textit{e.g.}, \textit{what impact does faith have on us?}). We present the detailed guidance for categorizing in Table~\ref{tab:guidance-granularity}.

\begin{figure}[h] 
    \centering 

    \begin{subfigure}[b]{0.32\textwidth} 
        \includegraphics[width=\textwidth]{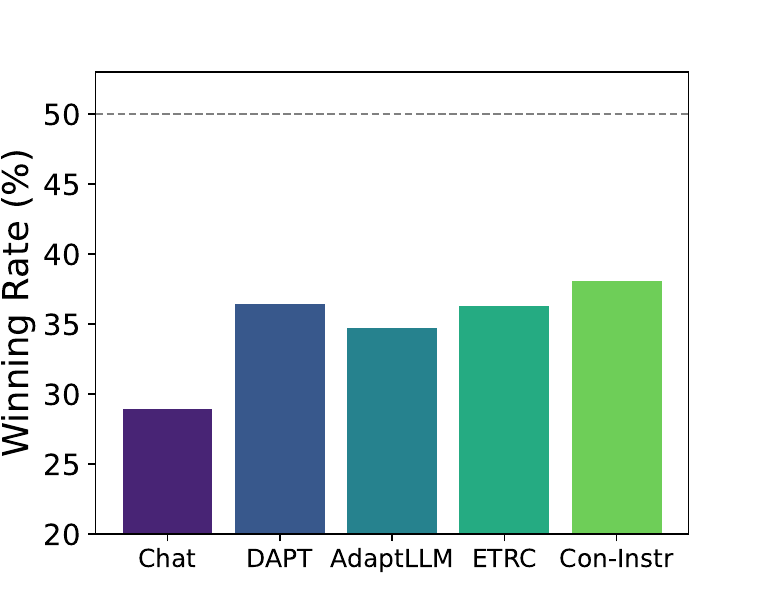} 
        \caption{Detail question}
        \label{fig:gpt4-judege-detail}
    \end{subfigure}
    \begin{subfigure}[b]{0.32\textwidth}
        \includegraphics[width=\textwidth]{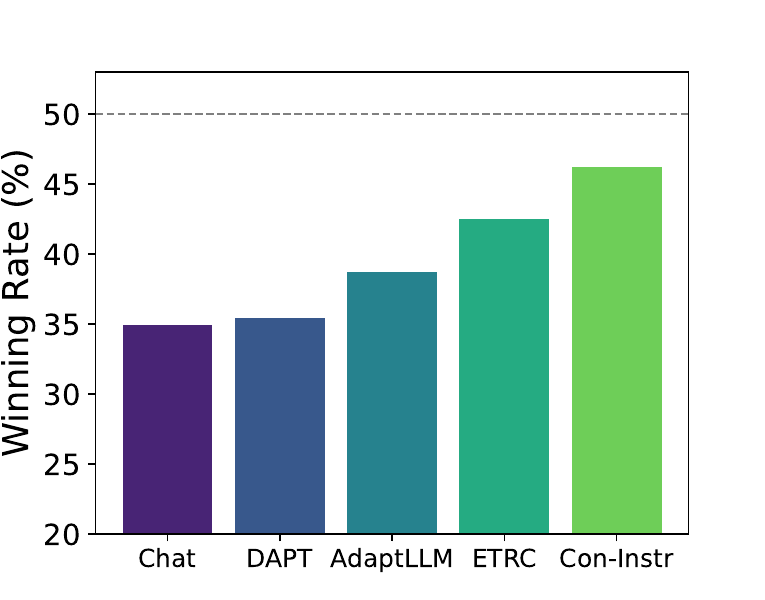}
        \caption{Concept question}
        \label{fig:gpt4-judege-concept}
    \end{subfigure}
    \begin{subfigure}[b]{0.32\textwidth}
        \includegraphics[width=\textwidth]{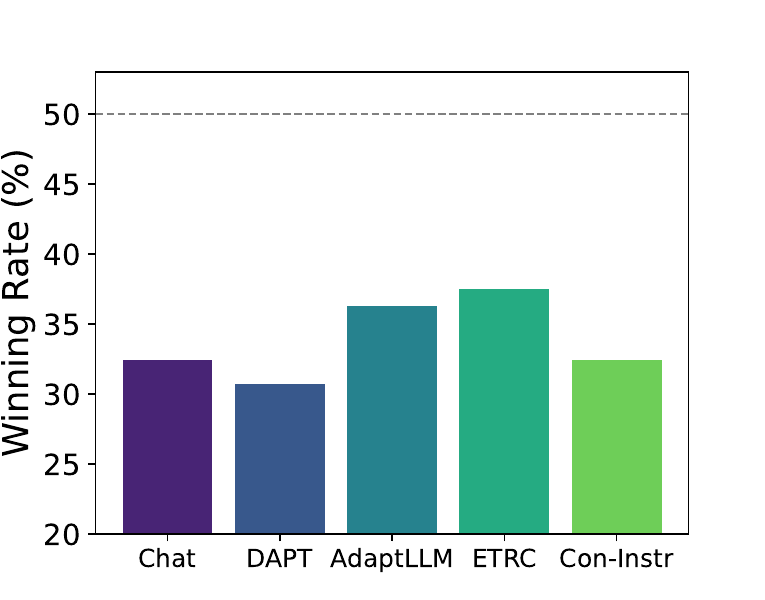}
        \caption{Macro question}
        \label{fig:gpt4-judege-macro}
    \end{subfigure}

    \caption{The results of GPT-4 judge on three levels of questions.}
    \label{fig:gpt4-judege}    
\end{figure}

For the reliability of the results, we extract relevant references to the questions in the dataset corpus to assist GPT-4 in making decisions. Then, we ask GPT-4 to compare the outputs generated by the two models, with the template shown in below text box. To mitigate the potential impact of position bias of LLMs, we implement a robust evaluation strategy that for each pair of outputs, we swap their positions and queried GPT-4 twice. In cases where the two responses were not consistent, we continued to inquire until we obtained a unanimous answer. The comparison results are shown in Figure~\ref{fig:gpt4-judege}.

\begin{tcolorbox}[breakable,colback=white!95!gray,colframe=gray!50!black,rounded corners,label={gpt4-judge-template}, title={GPT-4 Judge Template}]

Given a question and relevant reference, decide which one is better between answer 1 and answer 2.

\textbf{Question: }\{Question\}

\textbf{Reference: }\{Reference\}

\textbf{Answer 1: }\{Answer1\}

\textbf{Answer 2: }\{Answer2\}

\textbf{Your decision: }

\end{tcolorbox}

The performance comparisons are depicted in Figure~\ref{fig:gpt4-judege}. Note that since we employ a decisive, non-tie judgment, for a method to be considered superior, it must surpass the 50\% winning threshold. Through this rigorous comparison, it becomes apparent that our proposed method significantly outperforms the baseline methods across all three categories of question granularity. Notably, the winning rates for the baseline methods generally fall below 40\% in most instances, underscoring the effectiveness of our approach.

Among the baseline methods evaluated, ETRC and Context-Instruct exhibit relatively better performance in the domain of concept questions, with winning rates slightly exceeding 40\%. This performance is commendable when compared to other baseline methods. However, it's important to acknowledge that even these higher-performing baselines do not meet the 50\% threshold and exhibit weaknesses across the other two question granularities. This observation highlights a critical advantage of our method: its ability to maintain a balanced focus across different types of questions. Consequently, our approach not only excels in one particular area but also enhances performance across all three types of queries simultaneously.

The results from this comparison clearly illustrate the superior capability of our method in handling a diverse range of question complexities and types. This balanced focus is pivotal for developing systems that can adapt to varied informational needs, thereby improving the overall query response performance. Our findings suggest that our method could significantly contribute to advancements in improving the performance of fine-tuned models by offering a more versatile and effective approach to extracting multi-granularity query-response pairs from context.

\subsection{Computation Experiment}
\label{appx:computation-experiment}

In previous experiments, since we find that all methods spend much more time on final fine-tuning compared to the previous generation, we maintained that each method produced an equivalent number of query response pairs for the given contexts to balance computing resources. Nonetheless, when examining methods like ETRC, Context-Instruct, and our \methodname{} that need LLM calls, we observed variations in the computational time required to generate each query response data pair. In this evaluation, we control the LLM computation to be the same, all gave $80$ A100 GPU hours for running the workflow to generate data from the \datasetname{} dataset and then compare the fine-tuned model using GPT-4 judge on \datasetname{} test set. The results are shown in Table~\ref{tab:computation}.

\begin{table}[h]
  \centering
  \begin{tabular}{c|c||c|c}\toprule 
     & \multicolumn{2}{|c|}{Winning Rate} &  \\\midrule
   \methodname{}  &  $64.5\%$ & $35.5\%$ & ETRC \\\midrule
   \methodname{}  &  $60.3\%$ & $39.7\%$ & Con-Instr \\\bottomrule
  \end{tabular}
  \vspace{0.2cm}
  \caption{The winning rates in computation experiment.}
  \label{tab:computation}
\end{table}

In our comparative analysis, we discovered that our \methodname{} consistently achieved higher winning rates when set against two other LLM-based, context-driven query extraction methodologies. This finding is particularly significant as it validates the efficacy of our method in yielding positive outcomes even with a reduced quantity of query-response pairs. The cornerstone of this enhanced performance lies in the superior quality and diversity of the data pairs generated by our method. Unlike conventional approaches that may produce voluminous but redundant data, our method focuses on creating data pairs that are both essential and varied. This strategic approach to data generation ensures that \methodname{} operates efficiently, making the most of every data pair to contribute meaningfully to the fine-tuning of LLMs. As a result, \methodname{} stands out as a highly data-efficient method, capable of fine-tuning LLMs with less generated data to achieve better performance. This advantage is especially crucial in scenarios where access to large amounts of the corpus or the computation resource is constrained. By eliminating redundancy and emphasizing the importance of quality and diversity, \methodname{} paves the way for more effective and efficient context-based SFT data generation processes, ultimately leading to enhanced model performance even with limited sources.






\subsection{Training Phase}
\label{appx:training-phase}

\begin{figure}[h]
    \centering
    \includegraphics[width=0.7\linewidth]{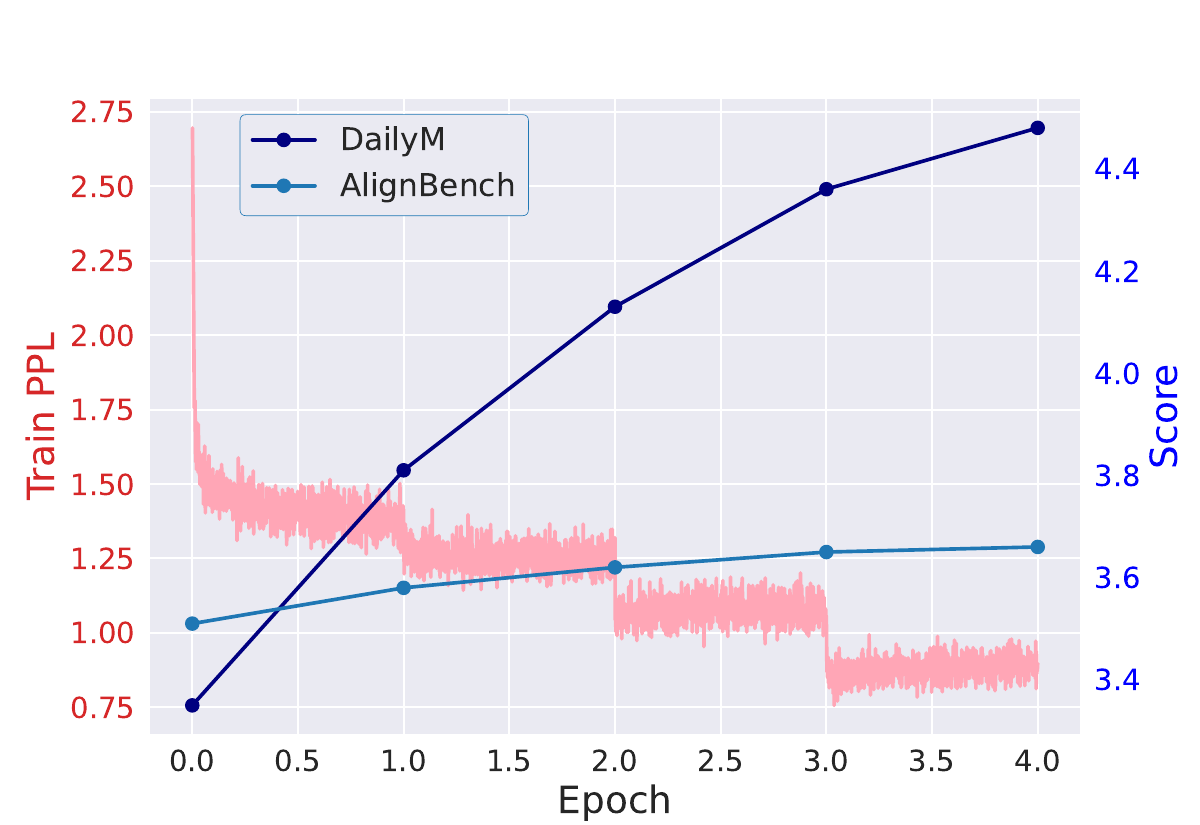}
    \caption{The training loss and human evaluation results during training phase.}
    \label{fig:training-phase}
\end{figure}

We present the loss curve training on the generated \sftdata{} in Figure~\ref{fig:training-phase}. From the graph, it is evident that the training loss decreases gradually in general throughout the training phase, with a notable sharp decline at the outset. An interesting observation is that the training loss appears to plateau within epochs from Epoch 2 onwards, yet we observe sudden drops in loss at the boundaries between two consecutive epochs. This pattern strongly signals that our training dataset is characterized by extremely low similarity and exceptionally high diversity, meaning that training on one segment of data does not have an impact on the loss associated with another segment. Conversely, if the dataset were more homogeneous, for instance, by including duplicates of certain data points, we would likely observe some decrease in the loss within epochs, just like the drop in two consecutive epochs in our curve. This phenomenon also indicates that the knowledge encapsulated in our responses is remarkably diverse too, covering a broad spectrum of information in the extracted domain corpus with minimal data usage, which is important for improving the performance of fine-tuned LLMs according to existing research~\cite{gunasekar2023textbooks,li2023textbooks}. 

We conduct a comprehensive human evaluation at each checkpoint during model training, and the overall satisfaction scores are presented in Figure~\ref{fig:training-phase}. Our results show that, on the \datasetname{} test set, the evaluation scores exhibit a rapid and steady increase as training progresses. Furthermore, to verify that our training does not compromise the model's general conversational abilities, we also employ AlignBench~\cite{liu2023alignbench}, a widely used benchmark that assesses the chat models abilities across multiple dimensions, including language understanding, question answering, writing, reasoning, and so on. Notably, thanks to the wide range of high-quality daily-focused articles provided by our \datasetname{} corpus, our model trained on the derived \sftdata{} not only preserves its original general abilities but also demonstrates a slight improvement, which also highlights the advantages of our methodology.

\subsection{Granularity Comparison}

In exploring the landscape of multi-granularity question generation, our methodology introduces a nuanced approach to constructing questions across different levels of granularity. We categorize questions into three distinct types based on their scope and depth: detail questions, concept questions, and macro questions, with the detailed guidance for categorizing shown in Table~\ref{tab:guidance-granularity}. This categorization allows us to systematically address the varying levels of knowledge Q\&A intends and users' interests.

\begin{figure}[h] 
    \centering 
    \begin{minipage}[t]{0.45\linewidth}
    \centering
    \begin{subfigure}[b]{0.7\textwidth} 
        \caption{\methodname{}}
        \includegraphics[width=\textwidth]{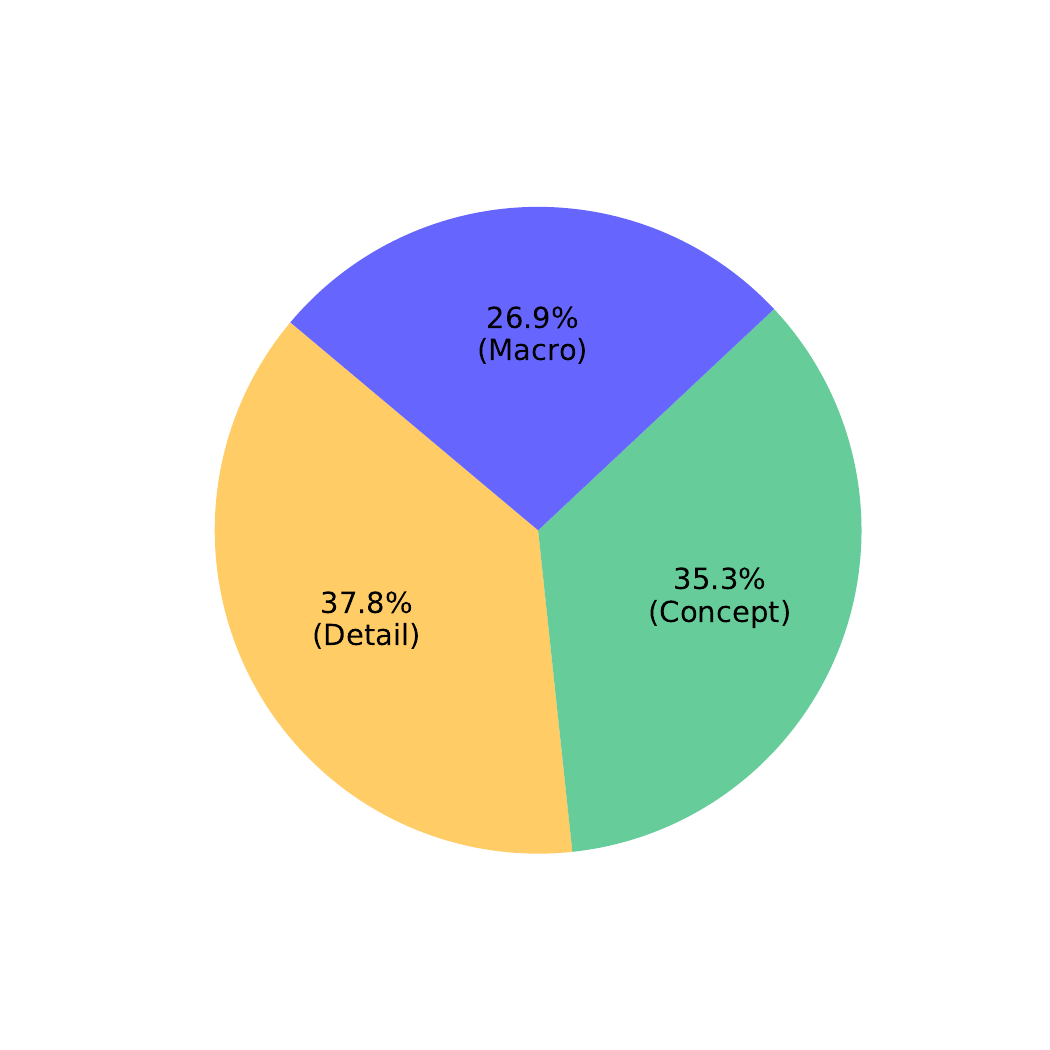} 
    \end{subfigure}
    \end{minipage}
    \hfill
    \begin{minipage}[t]{0.45\linewidth}
    \centering
    \begin{subfigure}[b]{0.7\textwidth}
        \caption{Context-Instruct}
        \includegraphics[width=\textwidth]{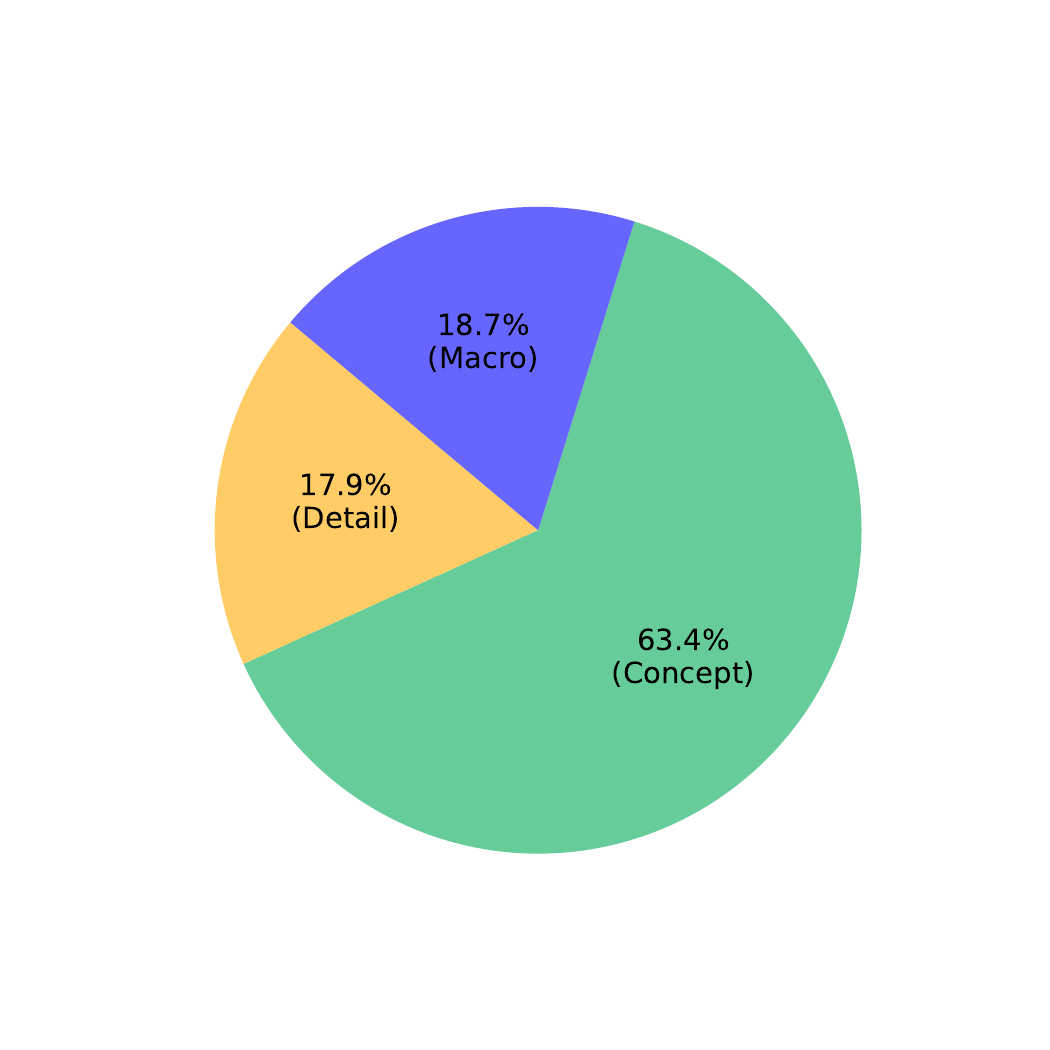}
    \end{subfigure}
    \end{minipage}
    \caption{The proportion of three levels of granularity questions generated by \methodname{} and Context-Instruct.}
    \label{fig:granularity-comparison}   
    
\end{figure}

The proportions of three types of questions are displayed in Figure~\ref{fig:granularity-comparison}. Our approach yields a balanced distribution of question types, with 37.8\% detail questions, 35.3\% concept questions, and 26.9\% macro questions. This distribution is indicative of our method's capacity to generate a diverse array of questions, catering to a broad spectrum of informational needs and cognitive processes. The balance ensures that the LLM is not only able to obtain specific information (detail questions) but is also able to grasp broader concepts (concept questions) and understand overarching themes or ideas (macro questions).

In contrast, when compared to the Context-Instruct method, which produces questions with a distribution of 17.9\% detail questions, 63.4\% concept questions, and 18.7\% macro questions, the advantages of our approach become evident. The Context-Instruct method exhibits a pronounced emphasis on concept questions, which, while valuable, suggests a potential oversight of the importance of detail and macro questions. Such distribution may limit the method's effectiveness in scenarios where a more detailed understanding or a broader overview is essential.

Additionally, combined with the previous analysis of the CST process, our method is superior in constructing a multi-granularity question set that is not only balanced but also versatile. By ensuring a more equitable distribution of question types, our approach enhances the comprehensiveness and depth of information exploration. It facilitates a more holistic understanding of content, accommodating a wider range of learning styles and information-seeking behaviors. This versatility is paramount in improving the fine-tuned LLMs' ability to tackle a wider array of questions and applications.

\begin{table}[h]
\centering
\caption{The guidance to categorize each granularity of questions.}
\label{tab:guidance-granularity}
\begin{tabular}{>{\bfseries}m{1.6cm} m{11.4cm}}
\toprule
Category & \textbf{Description and Examples} \\\midrule

Detail Questions & \textbf{Description:} Detail questions ask for specific information or facts about a particular aspect of a broader topic. These questions are often precise and seek exact answers.\\
& \textbf{Tips for Identification}: 
\begin{itemize}[leftmargin=0.8cm]
    \item Look for questions asking for “how,” “what,” “where,” “when,” or “who” in a specific context.
    \item These questions usually focus on a narrow aspect rather than the whole topic. 
\end{itemize} \\
&\textbf{Examples:}
\begin{itemize}[leftmargin=0.8cm]
    \item How deep does the abortion needle penetrate?
    \item What is the boiling point of mercury?
    \item Who was the first person to climb Mount Everest?
\end{itemize} \\
\midrule
Concept Questions & 
\textbf{Description:} Concept questions explore the understanding, differences, or definitions of ideas, theories, or terminologies. They are more about the “why” or “what is” than about specific details.\\
& \textbf{Tips for Identification}: 
\begin{itemize}[leftmargin=0.8cm]
    \item These questions often ask for explanations, comparisons, or definitions.
    \item They seek to clarify concepts or understand the distinctions between them.
\end{itemize} \\
& \textbf{Examples:}
\begin{itemize}[leftmargin=0.8cm]
    \item What is the difference between emergencies and crises?
    \item How do you differentiate between classical and operant conditioning?
    \item What does the term 'biodiversity' encompass?
\end{itemize} \\
\midrule
Macro Questions & 
\textbf{Description:} Macro questions address broad themes, trends, or impacts. They are expansive and consider the bigger picture, often relating to societal, global, or philosophical inquiries.\\
& \textbf{Tips for Identification}: 
\begin{itemize}[leftmargin=0.8cm]
    \item Look for questions that ask about “impact,” “influence,” “role,” or “importance.”
    \item These questions are overarching and not limited to specific instances.
\end{itemize} \\
&\textbf{Examples:}
\begin{itemize}[leftmargin=0.8cm]
    \item What impact does faith have on us?
    \item How does climate change affect global agriculture?
    \item What is the role of technology in shaping modern education?
\end{itemize} \\
\bottomrule
\end{tabular}
\end{table}

\clearpage

\section*{NeurIPS Paper Checklist}

\begin{enumerate}

\item {\bf Claims}
    \item[] Question: Do the main claims made in the abstract and introduction accurately reflect the paper's contributions and scope?
    \item[] Answer: \answerYes{}
    \item[] Justification: The main claims made in the abstract and introduction accurately reflect the paper's contributions and scope.
    
    \item[] Guidelines:
    \begin{itemize}
        \item The answer NA means that the abstract and introduction do not include the claims made in the paper.
        \item The abstract and/or introduction should clearly state the claims made, including the contributions made in the paper and important assumptions and limitations. A No or NA answer to this question will not be perceived well by the reviewers. 
        \item The claims made should match theoretical and experimental results, and reflect how much the results can be expected to generalize to other settings. 
        \item It is fine to include aspirational goals as motivation as long as it is clear that these goals are not attained by the paper. 
    \end{itemize}

\item {\bf Limitations}
    \item[] Question: Does the paper discuss the limitations of the work performed by the authors?
    \item[] Answer: \answerYes{} 
    \item[] Justification: We discuss the limitations of the work in Appendix~\ref{appx:limitations}.
    \item[] Guidelines:
    \begin{itemize}
        \item The answer NA means that the paper has no limitation while the answer No means that the paper has limitations, but those are not discussed in the paper. 
        \item The authors are encouraged to create a separate "Limitations" section in their paper.
        \item The paper should point out any strong assumptions and how robust the results are to violations of these assumptions (e.g., independence assumptions, noiseless settings, model well-specification, asymptotic approximations only holding locally). The authors should reflect on how these assumptions might be violated in practice and what the implications would be.
        \item The authors should reflect on the scope of the claims made, e.g., if the approach was only tested on a few datasets or with a few runs. In general, empirical results often depend on implicit assumptions, which should be articulated.
        \item The authors should reflect on the factors that influence the performance of the approach. For example, a facial recognition algorithm may perform poorly when image resolution is low or images are taken in low lighting. Or a speech-to-text system might not be used reliably to provide closed captions for online lectures because it fails to handle technical jargon.
        \item The authors should discuss the computational efficiency of the proposed algorithms and how they scale with dataset size.
        \item If applicable, the authors should discuss possible limitations of their approach to address problems of privacy and fairness.
        \item While the authors might fear that complete honesty about limitations might be used by reviewers as grounds for rejection, a worse outcome might be that reviewers discover limitations that aren't acknowledged in the paper. The authors should use their best judgment and recognize that individual actions in favor of transparency play an important role in developing norms that preserve the integrity of the community. Reviewers will be specifically instructed to not penalize honesty concerning limitations.
    \end{itemize}

\item {\bf Theory Assumptions and Proofs}
    \item[] Question: For each theoretical result, does the paper provide the full set of assumptions and a complete (and correct) proof?
    \item[] Answer: \answerYes{} 
    \item[] Justification: We have a proof on linear relationship and we provide the full set of assumptions and a complete and correct proof in Appendix~\ref{appx:CST-proof}.
    \item[] Guidelines:
    \begin{itemize}
        \item The answer NA means that the paper does not include theoretical results. 
        \item All the theorems, formulas, and proofs in the paper should be numbered and cross-referenced.
        \item All assumptions should be clearly stated or referenced in the statement of any theorems.
        \item The proofs can either appear in the main paper or the supplemental material, but if they appear in the supplemental material, the authors are encouraged to provide a short proof sketch to provide intuition. 
        \item Inversely, any informal proof provided in the core of the paper should be complemented by formal proofs provided in appendix or supplemental material.
        \item Theorems and Lemmas that the proof relies upon should be properly referenced. 
    \end{itemize}

    \item {\bf Experimental Result Reproducibility}
    \item[] Question: Does the paper fully disclose all the information needed to reproduce the main experimental results of the paper to the extent that it affects the main claims and/or conclusions of the paper (regardless of whether the code and data are provided or not)?
    \item[] Answer: \answerYes{} 
    \item[] Justification: We disclose our implementation details including all crucial hyperparameters, training configurations, and the used models and datasets in Appendix~\ref{appx:implementation-details}, and our method is elaborated in detail in Section~\ref{sec:method} and Appendix~\ref{appx:method-details} attached with specific prompts, which will be sufficient to reproduce the main experimental results of the paper to the extent that it affects the main claims and/or conclusions of the paper.
    \item[] Guidelines:
    \begin{itemize}
        \item The answer NA means that the paper does not include experiments.
        \item If the paper includes experiments, a No answer to this question will not be perceived well by the reviewers: Making the paper reproducible is important, regardless of whether the code and data are provided or not.
        \item If the contribution is a dataset and/or model, the authors should describe the steps taken to make their results reproducible or verifiable. 
        \item Depending on the contribution, reproducibility can be accomplished in various ways. For example, if the contribution is a novel architecture, describing the architecture fully might suffice, or if the contribution is a specific model and empirical evaluation, it may be necessary to either make it possible for others to replicate the model with the same dataset, or provide access to the model. In general. releasing code and data is often one good way to accomplish this, but reproducibility can also be provided via detailed instructions for how to replicate the results, access to a hosted model (e.g., in the case of a large language model), releasing of a model checkpoint, or other means that are appropriate to the research performed.
        \item While NeurIPS does not require releasing code, the conference does require all submissions to provide some reasonable avenue for reproducibility, which may depend on the nature of the contribution. For example
        \begin{enumerate}
            \item If the contribution is primarily a new algorithm, the paper should make it clear how to reproduce that algorithm.
            \item If the contribution is primarily a new model architecture, the paper should describe the architecture clearly and fully.
            \item If the contribution is a new model (e.g., a large language model), then there should either be a way to access this model for reproducing the results or a way to reproduce the model (e.g., with an open-source dataset or instructions for how to construct the dataset).
            \item We recognize that reproducibility may be tricky in some cases, in which case authors are welcome to describe the particular way they provide for reproducibility. In the case of closed-source models, it may be that access to the model is limited in some way (e.g., to registered users), but it should be possible for other researchers to have some path to reproducing or verifying the results.
        \end{enumerate}
    \end{itemize}

\item {\bf Open access to data and code}
    \item[] Question: Does the paper provide open access to the data and code, with sufficient instructions to faithfully reproduce the main experimental results, as described in supplemental material?
    \item[] Answer: \answerYes{} 
    \item[] Justification: We provide the links to the used models and datasets in Appendix~\ref{appx:assets} and provide our code of~\methodname{}, proposed dataset~\datasetname{}, generated SFT dataset~\sftdata{} and fine-tuned model~\modelname{} with sufficient instructions to faithfully reproduce the main experimental results in our GitHub repository~\githublink{}.
    
    \item[] Guidelines:
    \begin{itemize}
        \item The answer NA means that paper does not include experiments requiring code.
        \item Please see the NeurIPS code and data submission guidelines (\url{https://nips.cc/public/guides/CodeSubmissionPolicy}) for more details.
        \item While we encourage the release of code and data, we understand that this might not be possible, so “No” is an acceptable answer. Papers cannot be rejected simply for not including code, unless this is central to the contribution (e.g., for a new open-source benchmark).
        \item The instructions should contain the exact command and environment needed to run to reproduce the results. See the NeurIPS code and data submission guidelines (\url{https://nips.cc/public/guides/CodeSubmissionPolicy}) for more details.
        \item The authors should provide instructions on data access and preparation, including how to access the raw data, preprocessed data, intermediate data, and generated data, etc.
        \item The authors should provide scripts to reproduce all experimental results for the new proposed method and baselines. If only a subset of experiments are reproducible, they should state which ones are omitted from the script and why.
        \item At submission time, to preserve anonymity, the authors should release anonymized versions (if applicable).
        \item Providing as much information as possible in supplemental material (appended to the paper) is recommended, but including URLs to data and code is permitted.
    \end{itemize}

\item {\bf Experimental Setting/Details}
    \item[] Question: Does the paper specify all the training and test details (e.g., data splits, hyperparameters, how they were chosen, type of optimizer, etc.) necessary to understand the results?
    \item[] Answer: \answerYes{} 
    \item[] Justification: All training and testing details of our work can be found in Appendix~\ref{appx:implementation-details} and Section~\ref{sec:evaluations}
    \item[] Guidelines:
    \begin{itemize}
        \item The answer NA means that the paper does not include experiments.
        \item The experimental setting should be presented in the core of the paper to a level of detail that is necessary to appreciate the results and make sense of them.
        \item The full details can be provided either with the code, in appendix, or as supplemental material.
    \end{itemize}

\item {\bf Experiment Statistical Significance}
    \item[] Question: Does the paper report error bars suitably and correctly defined or other appropriate information about the statistical significance of the experiments?
    \item[] Answer: \answerYes{} 
    \item[] Justification: We report error bars suitably and correctly defined in Table~\ref{tab:result-auto-eval}.
    \item[] Guidelines:
    \begin{itemize}
        \item The answer NA means that the paper does not include experiments.
        \item The authors should answer "Yes" if the results are accompanied by error bars, confidence intervals, or statistical significance tests, at least for the experiments that support the main claims of the paper.
        \item The factors of variability that the error bars are capturing should be clearly stated (for example, train/test split, initialization, random drawing of some parameter, or overall run with given experimental conditions).
        \item The method for calculating the error bars should be explained (closed form formula, call to a library function, bootstrap, etc.)
        \item The assumptions made should be given (e.g., Normally distributed errors).
        \item It should be clear whether the error bar is the standard deviation or the standard error of the mean.
        \item It is OK to report 1-sigma error bars, but one should state it. The authors should preferably report a 2-sigma error bar than state that they have a 96\% CI, if the hypothesis of Normality of errors is not verified.
        \item For asymmetric distributions, the authors should be careful not to show in tables or figures symmetric error bars that would yield results that are out of range (e.g. negative error rates).
        \item If error bars are reported in tables or plots, The authors should explain in the text how they were calculated and reference the corresponding figures or tables in the text.
    \end{itemize}

\item {\bf Experiments Compute Resources}
    \item[] Question: For each experiment, does the paper provide sufficient information on the computer resources (type of compute workers, memory, time of execution) needed to reproduce the experiments?
    \item[] Answer: \answerYes{} 
    \item[] Justification: We provide sufficient information on the computer resources (type of compute workers, memory, time of execution) needed to reproduce the experiments in Appendix~\ref{appx:implementation-details}~and~\ref{appx:computation-experiment}.
    \item[] Guidelines:
    \begin{itemize}
        \item The answer NA means that the paper does not include experiments.
        \item The paper should indicate the type of compute workers CPU or GPU, internal cluster, or cloud provider, including relevant memory and storage.
        \item The paper should provide the amount of compute required for each of the individual experimental runs as well as estimate the total compute. 
        \item The paper should disclose whether the full research project required more compute than the experiments reported in the paper (e.g., preliminary or failed experiments that didn't make it into the paper). 
    \end{itemize}
    
\item {\bf Code Of Ethics}
    \item[] Question: Does the research conducted in the paper conform, in every respect, with the NeurIPS Code of Ethics \url{https://neurips.cc/public/EthicsGuidelines}?
    \item[] Answer: \answerYes{} 
    \item[] Justification: The research conducted in the paper conform, in every respect, with the NeurIPS Code of Ethics.
    \item[] Guidelines:
    \begin{itemize}
        \item The answer NA means that the authors have not reviewed the NeurIPS Code of Ethics.
        \item If the authors answer No, they should explain the special circumstances that require a deviation from the Code of Ethics.
        \item The authors should make sure to preserve anonymity (e.g., if there is a special consideration due to laws or regulations in their jurisdiction).
    \end{itemize}

\item {\bf Broader Impacts}
    \item[] Question: Does the paper discuss both potential positive societal impacts and negative societal impacts of the work performed?
    \item[] Answer: \answerYes{} 
    \item[] Justification: We discuss both potential positive societal impacts and negative societal impacts of the work in Appendix~\ref{appx:boarder-impacts}.
    \item[] Guidelines:
    \begin{itemize}
        \item The answer NA means that there is no societal impact of the work performed.
        \item If the authors answer NA or No, they should explain why their work has no societal impact or why the paper does not address societal impact.
        \item Examples of negative societal impacts include potential malicious or unintended uses (e.g., disinformation, generating fake profiles, surveillance), fairness considerations (e.g., deployment of technologies that could make decisions that unfairly impact specific groups), privacy considerations, and security considerations.
        \item The conference expects that many papers will be foundational research and not tied to particular applications, let alone deployments. However, if there is a direct path to any negative applications, the authors should point it out. For example, it is legitimate to point out that an improvement in the quality of generative models could be used to generate deepfakes for disinformation. On the other hand, it is not needed to point out that a generic algorithm for optimizing neural networks could enable people to train models that generate Deepfakes faster.
        \item The authors should consider possible harms that could arise when the technology is being used as intended and functioning correctly, harms that could arise when the technology is being used as intended but gives incorrect results, and harms following from (intentional or unintentional) misuse of the technology.
        \item If there are negative societal impacts, the authors could also discuss possible mitigation strategies (e.g., gated release of models, providing defenses in addition to attacks, mechanisms for monitoring misuse, mechanisms to monitor how a system learns from feedback over time, improving the efficiency and accessibility of ML).
    \end{itemize}
    
\item {\bf Safeguards}
    \item[] Question: Does the paper describe safeguards that have been put in place for responsible release of data or models that have a high risk for misuse (e.g., pretrained language models, image generators, or scraped datasets)?
    \item[] Answer: \answerNA{} 
    \item[] Justification: Our released model is fine-tuned from the existing open-source model with SFT data derived from high-quality magazines closely related to daily life, which contains no harmful or misleading information and does not have such risks of misuse.
    \item[] Guidelines:
    \begin{itemize}
        \item The answer NA means that the paper poses no such risks.
        \item Released models that have a high risk for misuse or dual-use should be released with necessary safeguards to allow for controlled use of the model, for example by requiring that users adhere to usage guidelines or restrictions to access the model or implementing safety filters. 
        \item Datasets that have been scraped from the Internet could pose safety risks. The authors should describe how they avoided releasing unsafe images.
        \item We recognize that providing effective safeguards is challenging, and many papers do not require this, but we encourage authors to take this into account and make a best faith effort.
    \end{itemize}

\item {\bf Licenses for existing assets}
    \item[] Question: Are the creators or original owners of assets (e.g., code, data, models), used in the paper, properly credited and are the license and terms of use explicitly mentioned and properly respected?
    \item[] Answer: \answerYes{} 
    \item[] Justification: We conform to the license and terms of all used assets, as mentioned in Appendix~\ref{appx:assets}.
    \item[] Guidelines:
    \begin{itemize}
        \item The answer NA means that the paper does not use existing assets.
        \item The authors should cite the original paper that produced the code package or dataset.
        \item The authors should state which version of the asset is used and, if possible, include a URL.
        \item The name of the license (e.g., CC-BY 4.0) should be included for each asset.
        \item For scraped data from a particular source (e.g., website), the copyright and terms of service of that source should be provided.
        \item If assets are released, the license, copyright information, and terms of use in the package should be provided. For popular datasets, \url{paperswithcode.com/datasets} has curated licenses for some datasets. Their licensing guide can help determine the license of a dataset.
        \item For existing datasets that are re-packaged, both the original license and the license of the derived asset (if it has changed) should be provided.
        \item If this information is not available online, the authors are encouraged to reach out to the asset's creators.
    \end{itemize}

\item {\bf New Assets}
    \item[] Question: Are new assets introduced in the paper well documented and is the documentation provided alongside the assets?
    \item[] Answer: \answerYes{} 
    \item[] Justification: We provide the construction details of the assets in Section~\ref{sec:human-evaluation}, and the documentations are provided alongside the assets at \githublink{}.
    \item[] Guidelines:
    \begin{itemize}
        \item The answer NA means that the paper does not release new assets.
        \item Researchers should communicate the details of the dataset/code/model as part of their submissions via structured templates. This includes details about training, license, limitations, etc. 
        \item The paper should discuss whether and how consent was obtained from people whose asset is used.
        \item At submission time, remember to anonymize your assets (if applicable). You can either create an anonymized URL or include an anonymized zip file.
    \end{itemize}

\item {\bf Crowdsourcing and Research with Human Subjects}
    \item[] Question: For crowdsourcing experiments and research with human subjects, does the paper include the full text of instructions given to participants and screenshots, if applicable, as well as details about compensation (if any)? 
    \item[] Answer: \answerYes{} 
    \item[] Justification: We include the detailed human evaluation metrics in Section~\ref{sec:human-eval-metrics} and the full text of instructions in Appendix~\ref{appx:guidance}.
    \item[] Guidelines:
    \begin{itemize}
        \item The answer NA means that the paper does not involve crowdsourcing nor research with human subjects.
        \item Including this information in the supplemental material is fine, but if the main contribution of the paper involves human subjects, then as much detail as possible should be included in the main paper. 
        \item According to the NeurIPS Code of Ethics, workers involved in data collection, curation, or other labor should be paid at least the minimum wage in the country of the data collector. 
    \end{itemize}

\item {\bf Institutional Review Board (IRB) Approvals or Equivalent for Research with Human Subjects}
    \item[] Question: Does the paper describe potential risks incurred by study participants, whether such risks were disclosed to the subjects, and whether Institutional Review Board (IRB) approvals (or an equivalent approval/review based on the requirements of your country or institution) were obtained?
    \item[] Answer: \answerNA{} 
    \item[] Justification: Not human subjects research.
    \item[] Guidelines:
    \begin{itemize}
        \item The answer NA means that the paper does not involve crowdsourcing nor research with human subjects.
        \item Depending on the country in which research is conducted, IRB approval (or equivalent) may be required for any human subjects research. If you obtained IRB approval, you should clearly state this in the paper. 
        \item We recognize that the procedures for this may vary significantly between institutions and locations, and we expect authors to adhere to the NeurIPS Code of Ethics and the guidelines for their institution. 
        \item For initial submissions, do not include any information that would break anonymity (if applicable), such as the institution conducting the review.
    \end{itemize}

\end{enumerate}


\end{document}